\documentclass[11pt]{article}

\usepackage[preprint]{acl}

\usepackage[most]{tcolorbox}
\tcbuselibrary{breakable}
\usepackage{capt-of}
\usepackage{enumitem}
\usepackage{times}
\usepackage{latexsym}
\usepackage{algorithm}
\usepackage{algpseudocode}
\usepackage{amsmath}
\usepackage{multirow}
\usepackage[T1]{fontenc}
\usepackage{booktabs}
\usepackage[table]{xcolor}

\usepackage{xurl}
\usepackage{hyperref}

\usepackage{enumitem}
\setlist{topsep=2pt, itemsep=1pt, parsep=0pt}

\usepackage{graphicx}
\usepackage{subcaption}

\usepackage[utf8]{inputenc}

\usepackage{microtype}

\usepackage{tcolorbox}
\tcbuselibrary{breakable}

\usepackage{inconsolata}
\usepackage{enumitem}
\usepackage{amsmath}

\usepackage{graphicx}

\title{DynaDebate: Breaking Homogeneity in Multi-Agent Debate with Dynamic Path Generation}

\author{
Zhenghao Li\textsuperscript{1},
jinyu Liu\textsuperscript{1},
Zhi Zheng\textsuperscript{1},
Wei Chen\textsuperscript{1},
Jielun Zhao\textsuperscript{3},
Yong Chen\textsuperscript{2},
Tong Xu\textsuperscript{1},
Enhong Chen\textsuperscript{1} \\
{\normalfont\textsuperscript{1}State Key Laboratory of Cognitive Intelligence, University of Science and Technology of China} \\
{\normalfont\textsuperscript{2}North Automatic Control Technology Institute} \\
{\normalfont\textsuperscript{3}North Automatic Control Technology Institute; Shenzhen Institute for Advanced Study, UESTC} \\
{\normalfont\ttfamily\small lizhenghao@mail.ustc.edu.cn, craden@mail.ustc.edu.cn,chenweicw@mail.ustc.edu.cn} \\
{\normalfont\ttfamily\small zhengzhi97@ustc.edu.cn, tongxu@ustc.edu.cn, cheneh@ustc.edu.cn} \\
{\normalfont\ttfamily\small 202512281035@std.uestc.edu.cn, chenyong1997@163.com}
}

\begin{document}
\maketitle

\begin{abstract}
Recent years have witnessed the rapid development of Large Language Model-based Multi-Agent Systems (MAS), which excel at collaborative decision-making and complex problem-solving. Researchers have further investigated Multi-Agent Debate (MAD) frameworks, which enhance the reasoning and collaboration capabilities of MAS through information exchange and debate among multiple agents. However, existing approaches often rely on unguided initialization, causing agents to adopt identical reasoning paths that lead to the same errors. As a result, effective debate among agents is hindered, and the final outcome frequently degenerates into simple majority voting. To solve the above problem, we introduce Dynamic Multi-Agent Debate (DynaDebate), which enhances the effectiveness of multi-agent debate through three key mechanisms: (1) Dynamic Path Generation and Allocation, which employs a dedicated Path Generation Agent to generate diverse and logical solution paths with adaptive redundancy; (2) Process-Centric Debate, which shifts the focus from surface-level outcome voting to rigorous step-by-step logic critique to ensure process correctness; (3) A Trigger-Based Verification Agent, which is activated upon disagreement and uses external tools to objectively resolve deadlocks. Experiments show that DynaDebate achieves superior or highly competitive performance across the majority of benchmarks\footnote{The code is at https://github.com/nwpuLee2021/brianstorm.}.
\end{abstract}

\section{Introduction}
\label{sec:intro}

LLM-based autonomous agents have demonstrated strong capabilities in planning, tool use, and multi-step reasoning \cite{yao2022react,wang2023voyager,shen2023hugginggpt}. However, individual agents still struggle with tasks requiring diverse perspectives and collaborative verification. To address this, the field has evolved into Multi-Agent Systems, which orchestrate multi-agent collaboration to combine diverse viewpoints and surpass individual model performance \cite{park2023generative,li2024survey}.

Within the realm of Multi-Agent Systems, Multi-Agent Debate \cite{liang2023encouraging,du2023improving,chan2023chateval} has emerged as a particularly effective paradigm. In this framework, multiple agents are introduced to discuss a shared question. Agents first generate independent responses and then engage in a structured debate, iteratively refining their contributions based on peer feedback. This approach has shown significant improvements in factuality and reasoning tasks.

\begin{figure}[t]
    \centering
    \includegraphics[width=0.95\linewidth]{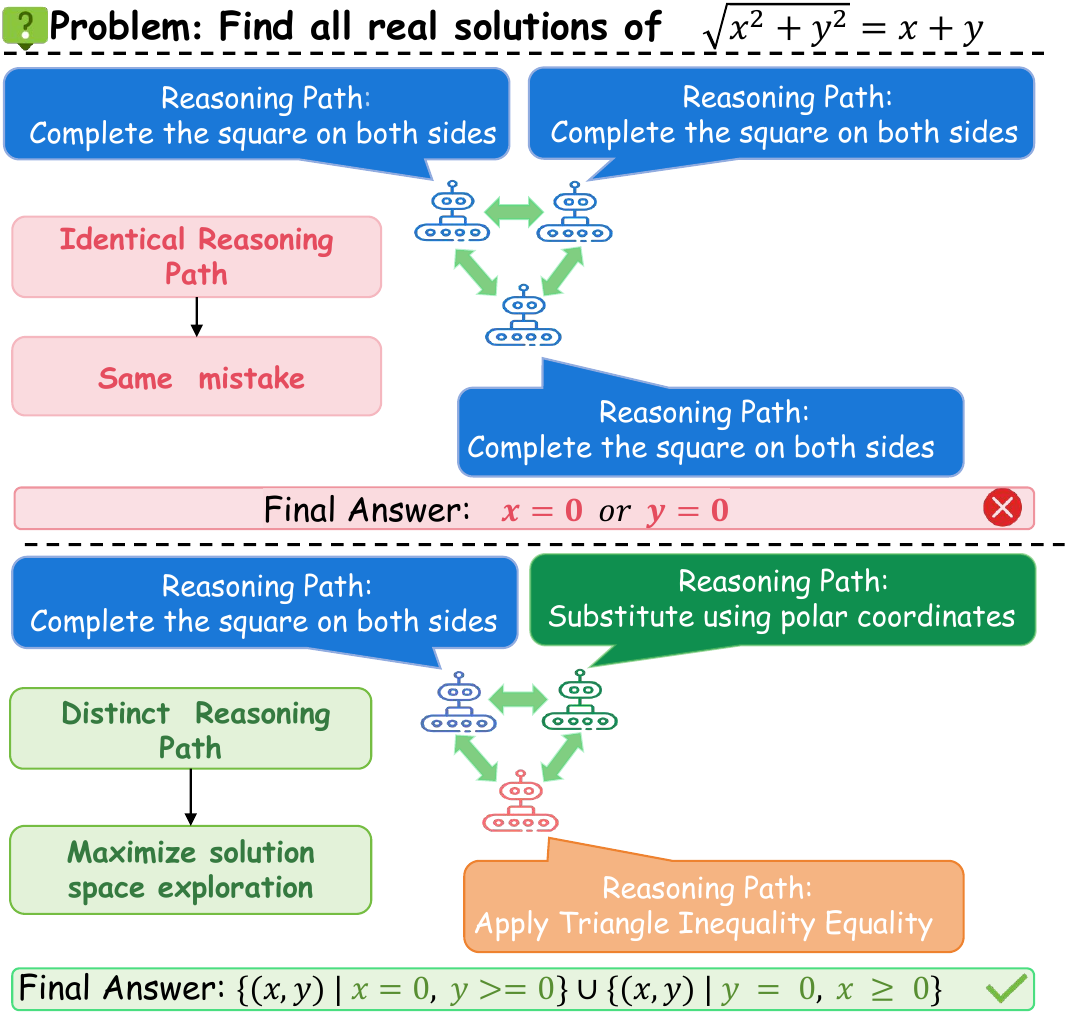}
    \caption{
    \textbf{Comparison of reasoning paths in the multi-agent debate phase.}
    (Top) Existing methods: Agents participating in the debate follow identical reasoning paths.
(Bottom) In the DynaDebate framework, agents are assigned distinct reasoning paths, ensuring diversity in reasoning paths.
    }
    \label{fig:motivation}
\end{figure}

However, a major challenge remains: due to model homogeneity and identical inherent mental sets \cite{liu2025breaking,zhang2025if,smit2023should}, agents often generate nearly identical responses during the debate, as illustrated in Figure \ref{fig:motivation} (top), causing the process to collapse to simple majority voting \cite{choi2025debate}.To address this, current approaches employ techniques such as static role-playing and reasoning strategy diversity. Despite these efforts, existing methods rely on unguided initialization before the debate phase. Consequently, agents frequently converge on the same reasoning path. When errors arise, they replicate the same mistake across agents, leaving the inter-agent debate unable to identify and correct the error. Furthermore, during the debate, agents often assess peer responses merely based on structural completeness or textual fluency. This superficial scrutiny exacerbates \textbf{blind conformity}: even when an agent holds the correct answer, it is susceptible to the erroneous reasoning of its peers, often abandoning its original stance in favor of an incorrect group consensus. Moreover, as debate rounds progress, erroneous views can accumulate and reinforce each other, making it increasingly difficult for agents to break free from a forming false consensus without the intervention of external, objective evidence.

To address these challenges, we introduce \textbf{DynaDebate}, a dynamic Multi-Agent Debate framework that introduces initialization heterogeneity, as illustrated in Figure~\ref{fig:motivation} (bottom). Distinct from existing approaches, our framework operates through a three-stage progressive pipeline. First, before the debate commences, a dedicated Path Generation Agent generates diverse and logically sound solution paths with adaptive redundancy. This mechanism effectively breaks initialization homogeneity, maximizing exploration of the solution space. Subsequently, during the debate, the framework enforces a process-centric protocol where agents audit specific reasoning steps rather than focusing solely on final answers. Specifically, agents perform a First-Principles Audit of the reasoning process instead of evaluating based on textual fluency. Finally, DynaDebate integrates a trigger-based Verification Agent. This agent invokes external tools to feed results back into the debate, helping agents resolve disagreements.

\begin{figure*}[t]
    \centering
    \includegraphics[width=0.95\textwidth]{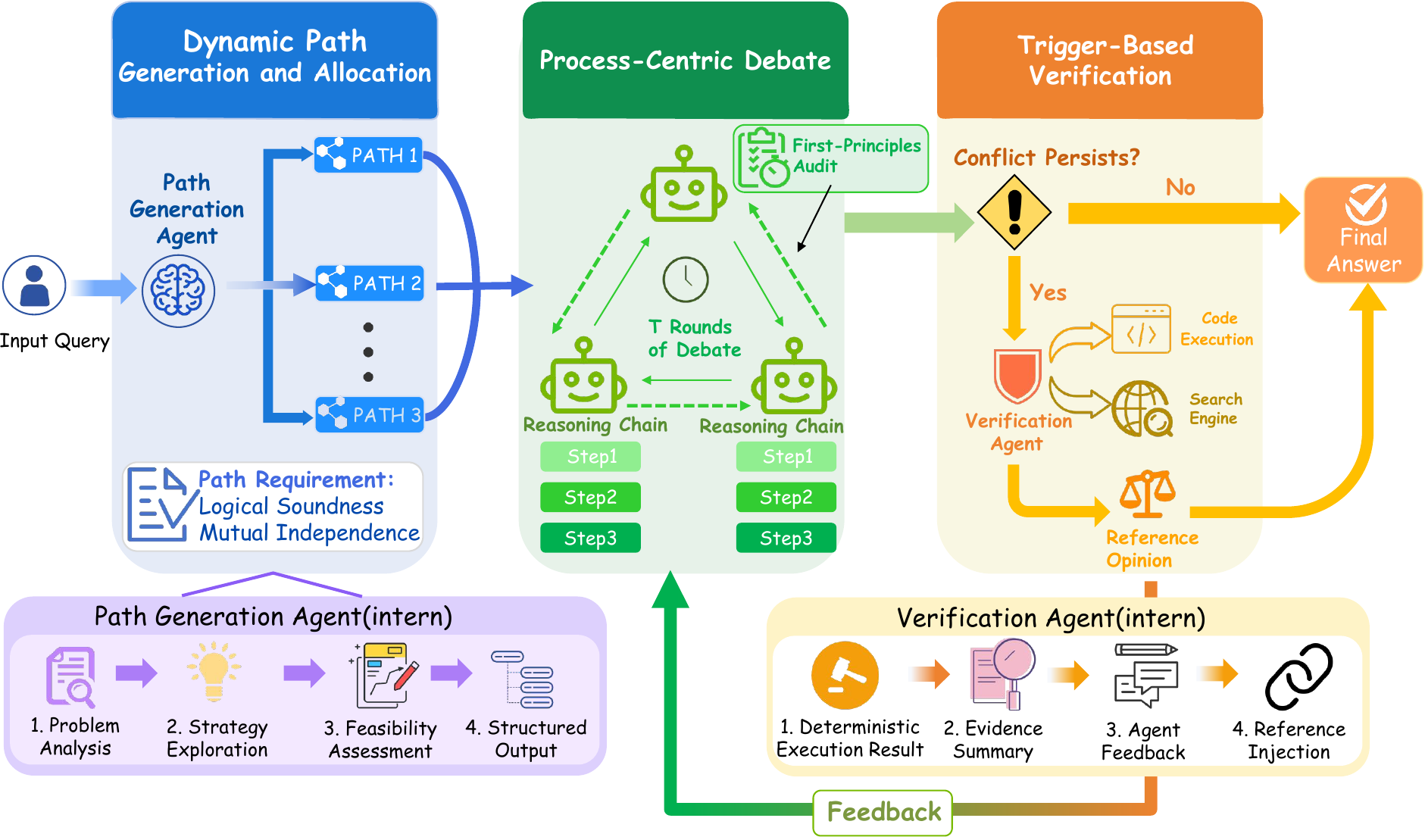}
    \caption{
        The overall framework of \textbf{DynaDebate}.
        The process operates in three phases:
        (1) \textbf{Dynamic Path Generation}: Initializes agents with diverse, logically sound paths to break homogeneity.
        (2) \textbf{Process-Centric Debate}: Agents perform a First-Principles Audit on peer reasoning steps.
        (3) \textbf{Trigger-Based Verification}: A Verification Agent integrates external tools to provide a reference for the debate and help resolve deadlocks.
    }
    \label{fig:framework}
\end{figure*}

Our contributions are summarized as follows:
\begin{enumerate}
    \item We introduce a dynamic path generation mechanism that enforces initialization heterogeneity, guiding agents to explore a broader solution space and thereby enhancing complex reasoning capability.
    \item We establish a rigorous process-centric protocol, integrated with a trigger-based verification mechanism to mitigate blind conformity and to ensure that consensus is firmly grounded in objective facts.
    \item Experiments show that DynaDebate achieves superior or highly competitive performance over existing Multi-Agent Debate methods across diverse benchmarks.
\end{enumerate}

\section{Related Works}

\paragraph{Multi-agent Debate}

Proposed by MAD \cite{liang2023encouraging} and SoM \cite{du2023improving}, Multi-Agent Debate has demonstrated significant performance improvements across diverse reasoning benchmarks. ChatEval \cite{chan2023chateval} utilizes this interaction mechanism for autonomous evaluation. Some research focuses on optimizing communication topologies \cite{li2024improving} to reduce token consumption, while other work integrates confidence scores \cite{lin2025enhancing} to weigh agent contributions. Regarding heterogeneity, Heter-MAD \cite{zhang2025if} investigates the impact of diversity, and DMAD  \cite{liu2025breaking} employs diverse prompting strategies to foster divergent reasoning.

\paragraph{Tool-Augmented Reasoning and Verification}

Recent works augment LLMs with external utilities to overcome closed-loop limitations. Toolformer \cite{schick2023toolformer} and ART \cite{paranjape2023art} enable models to autonomously invoke tools for multi-step reasoning, while ToolLLM \cite{qin2023toolllm} scales this capability to master real-world APIs. In mathematical domains, Program of Thoughts (PoT) \cite{chen2022program} disentangles computation from reasoning, and Code-based Self-Verification \cite{zhou2023solving} leverages code execution to verify solutions. Focusing on robustness, CRITIC \cite{gou2023critic} interacts with tools to validate outputs, and VerifiAgent \cite{han2025verifiagent} proposes a unified verification framework for reasoning tasks.

\paragraph{Multi-path reasoning}

To mitigate the instability of single-chain reasoning, researchers have developed multi-path strategies that explore diverse solution spaces. Self-Consistency (CoT-SC) \cite{wang2022self} pioneered this direction by sampling multiple reasoning paths and applying majority voting to filter out random errors. Moving beyond independent sampling, Tree of Thoughts (ToT) \cite{yao2023tree} structures reasoning as a tree search, enabling lookahead and backtracking via algorithms like BFS or DFS. Graph of Thoughts (GoT) \cite{besta2024graph} further generalizes this into a graph topology, allowing for the aggregation and recurrence of thoughts.

\section{Method}
\label{sec:method}
\subsection{Framework Overview}
\label{sec:overview}

We propose \textbf{DynaDebate}, a multi-agent reasoning framework designed to break initialization homogeneity and ensure process correctness. Formally, given a query $q$, our objective is to derive the correct final answer $y$. The system consists of a set of agents $\mathcal{A} = \{a_1, a_2, \dots, a_N\}$, which interact through a three-stage workflow as illustrated in Figure \ref{fig:framework} (pseudocode in Appendix~\ref{app:pseudocode}).

\subsection{Dynamic Path Generation and Allocation}
\label{sec:path_gen}

Current frameworks often rely on unguided initialization, rendering agents susceptible to generating identical reasoning paths, making it difficult to identify and correct shared errors. To mitigate this, DynaDebate introduces a structured initialization phase comprising Path Generation and Adaptive Redundancy Allocation.

\textbf{Path Generation.}
Prior to the debate, the system utilizes a Path Generation Agent, denoted as $\Phi_{\text{gen}}$, to perform a problem-specific analysis. Given the query $q$, the agent generates a set of candidate reasoning paths $\mathcal{P}$:
\begin{equation}
\mathcal{P} = \Phi_{\text{gen}}(q) = \{p_1, p_2, \ldots, p_K\},
\end{equation}
where $1 \le K \le N$.

The generation process is strictly guided by two constraints:
\begin{enumerate}
    \item \textbf{Logical Soundness:} Each candidate path $p_k$ must be a logically sound method specifically tailored to solving $q$. Logically unsound paths would introduce additional noise into the debate process.
    \item \textbf{Mutual Independence:} To avoid high similarity in agent reasoning and ensure system heterogeneity, we require reasoning paths to be mutually independent. Specifically, for computational tasks, agents employ distinct mathematical methodologies; for information retrieval tasks, they access diverse knowledge sources; and for logical reasoning tasks, they construct unique evidential chains. This constraint forces agents to deviate from the standard response patterns inherent in their training data, thereby ensuring the initial diversity of the solution space.
\end{enumerate}

\textbf{Adaptive Redundancy Allocation.}
To handle cases where the number of valid paths $K$ varies, we employ a Round-Robin Assignment mechanism. The path assigned to agent $a_i$ is defined as:
\begin{equation}
    \text{Path}(a_i) = p_{(i-1 \pmod K) + 1.}
\end{equation}
This formulation implements adaptive redundancy, balancing exploration diversity with consistency verification.
Specifically, in Exploration Mode ($K \approx N$), when multiple distinct paths are feasible, agents are assigned unique paths to encourage the exploration of diverse solutions.
In contrast, under Consistency Check Mode ($K < N$), where valid paths are limited, the same path is assigned to multiple agents, leveraging LLM stochasticity to detect and filter sporadic errors.

\subsection{Process-Centric Debate}
\label{sec:debate}

Following the initialization phase, agents engage in a multi-round debate. Unlike existing debate frameworks that primarily focus on final answers to reach a consensus, DynaDebate enforces a process-centric protocol. This protocol mandates that agents explicitly detail their reasoning process, enabling a rigorous step-by-step audit of the logic presented by their peers.

\textbf{Path-Guided Execution.}
In the initial round ($t=1$), each agent $a_i$ generates a solution strictly adhering to its assigned path $\text{Path}(a_i)$. Agents are prompted to decompose the problem-solving into a sequence of atomic inference steps. Formally, let $R_{i,1}$ denote the reasoning chain generated by agent $a_i$. We define $R_{i,1}$ not as unstructured text, but as an ordered set of logical steps:
\begin{equation}
    R_{i,1} = \{z_i^{(1)}, z_i^{(2)}, \dots, z_i^{(L)}\} \sim \pi( \cdot \mid q, \text{Path}(a_i)),
\end{equation}
where each $z_i^{(k)}$ represents a distinct calculation or logical deduction derived from the preceding steps $z_i^{(<k)}$. This explicit decomposition ensures that the reasoning trajectory is transparent, laying the foundation for the subsequent verification phase.

\textbf{Step-Level Peer Review.}
In subsequent rounds ($t>1$), agents receive the solutions generated by their peers. The core innovation of DynaDebate is the shift in the verification target. Debating agents no longer place their focus on the final answers generated by peers, nor do they assess the correctness of the reasoning process based on structural completeness or textual fluency. Instead, they are prompted to conduct a \textbf{First-Principles Audit} on the inference steps.

Specifically, for a peer's reasoning chain $R_j$, agent $a_i$ performs a rigorous examination targeting both \textbf{intrinsic correctness} and \textbf{logical coherence}. This involves verifying whether each individual step $z_j^{(k)}$ contains calculation or factual errors, while simultaneously auditing the validity of the logical transition from step $z_j^{(k)}$ to $z_j^{(k+1)}$. Agent $a_i$ identifies the specific location where either a factual mistake or a derivation gap occurs, focusing its critique strictly on these identified flaws. This granular scrutiny forces agents to abandon superficial agreement and instead converge solely on logically sound derivations.

\subsection{Trigger-Based Verification }
\label{sec:verification}

Although process-centric peer review strengthens the scrutiny of peer responses, agents may still remain susceptible to blind conformity. Specifically, even when an agent holds the correct answer, it can be influenced by erroneous peer reasoning and may ultimately abandon its original stance to align with the incorrect consensus \cite{wynn2025talk}. To address this limitation, drawing inspiration from the verification paradigm in VerifiAgent~\cite{han2025verifiagent}, DynaDebate integrates a dedicated Trigger-Based Verification Agent, denoted as $\Phi_{\text{ver}}$. The activation of this agent is conditional, governed by a function $\text{Trigger}(H_t)$: at the end of each debate round, if the participating agents produce divergent final answers, $\Phi_{\text{ver}}$ is activated; conversely, if all agents reach unanimous agreement, the debate terminates early without proceeding to the next round, thereby reducing unnecessary token consumption.

Upon activation, the agent autonomously selects and invokes appropriate tools (e.g., a Python interpreter or a web search engine) to execute the command based on the problem context. Given the history $H_t$, the agent generates a verification command and executes it in an external environment to obtain a deterministic observation $o_t \leftarrow \Phi_{\text{ver}}(q, H_t)$. The Verification Agent integrates this result to formulate a reference opinion, updating the debate history:
\begin{equation}
    H_t \leftarrow H_t \cup \{o_t\}.
\end{equation}
This input serves as a critical reference for the subsequent debate round, enabling participating agents to objectively scrutinize the discrepancies between their own stances and the provided evidence.

\section{Experiments}
\label{sec:exp}

To evaluate the versatility of DynaDebate, we conducted experiments on three backbone models: the proprietary model GPT-4o-mini and the open-source models Qwen3-8B and Llama-3.3-70B-Instruct. To facilitate a fair comparison with existing MAD methods, we standardized the multi-agent configuration to involve three agents engaging in two rounds of debate. Detailed prompts are provided in Appendix~\ref{app:prompts}.

\subsection{Experimental Setup}
\label{sec:setup}

We report results on five benchmarks in the main table; GSM8K and Biography results appear in appendices due to their distinct evaluation protocols (Appendix~\ref{app:gsm8k} and Appendix~\ref{app:biography}, respectively). We compare DynaDebate against seven representative baselines; detailed descriptions of dataset statistics and baseline implementations are provided in Appendix~\ref{app:datasets} and Appendix~\ref{app:baselines}, respectively.

\noindent\textbf{Benchmarks.} We categorize the evaluation tasks as follows:
(1) Mathematical Reasoning: We utilize MATH500 (comprehensive subjects) \cite{lightman2023lets} and AIME 2024-2025 (challenging competitions) \cite{aime2024, aime2025} to test reasoning capabilities at varying difficulty levels. Results on the near-saturated GSM8K benchmark \cite{cobbe2021gsm8k} are presented separately in Appendix~\ref{app:gsm8k}.
(2) General Knowledge: We employ MMLU \cite{hendrycks2020measuring} to assess knowledge breadth across diverse disciplines.
(3) Expert Knowledge Reasoning: We use the GPQA Main subset~\cite{rein2023gpqa} (448 questions) to evaluate graduate-level scientific reasoning across biology, physics, and chemistry.
(4) Factual Correctness: We evaluate on the Biography benchmark~\cite{du2023improving} for hallucination mitigation; full results are in Appendix~\ref{app:biography}.

\noindent\textbf{Evaluation Metrics}
We employ standard Accuracy for all benchmarks: MATH500, AIME, MMLU, and GPQA. The detailed evaluation protocol is provided in Appendix~\ref{app:datasets}. All results are mean $\pm$ std.\ (\%) over three runs.

\noindent\textbf{Baselines.} We compare our framework against a comprehensive set of methods:
(1) Single-Agent Strategies: We include CoT (standard prompting) \cite{wei2022chain}, CoT-SC (majority voting) \cite{wang2022self}, Self-Refine (iterative feedback) \cite{madaan2023self}, and Self-Contrast (contrastive self-evaluation) \cite{selfcontrast2024}. 
(2) Multi-Agent Frameworks: We benchmark against MAD (persona-based debate) \cite{liang2023encouraging}, SoM (Society of Mind) \cite{du2023improving}, and DMAD \cite{liu2025breaking}.
To ensure a rigorous comparison, we further examine whether the observed performance gains stem solely from tool usage rather than from multi-agent collaboration. We demonstrate in Appendix~\ref{sec:appendix_tool_ablation} that DynaDebate outperforms single-agent baselines even when they are augmented with identical external tools.

\begin{table*}[t]
\centering
\resizebox{\textwidth}{!}{
\begin{tabular}{llccccc}
\toprule
\textbf{Model} & \textbf{Method} & \textbf{MATH500} & \textbf{AIME 2024} & \textbf{AIME 2025} & \textbf{MMLU} & \textbf{GPQA} \\
\midrule

\multirow{8}{*}{\textbf{Qwen3-8B}}
 & CoT              & $78.07 \pm 0.12$                  & $30.00 \pm 3.33$                  & $21.11 \pm 3.85$                  & \underline{$81.89 \pm 2.79$}       & $43.67 \pm 0.84$ \\
 & CoT-SC           & $72.87 \pm 7.57$                  & \underline{$32.22 \pm 5.09$}       & $17.78 \pm 1.92$                  & $79.22 \pm 0.39$                  & $43.01 \pm 2.39$ \\
 & Self-Refine      & $79.87 \pm 1.14$                  & $23.33 \pm 3.33$                  & $17.78 \pm 1.92$                  & $69.11 \pm 2.01$                  & $36.90 \pm 1.31$ \\
 & Self-Contrast    & $78.00 \pm 1.60$                  & $22.22 \pm 6.94$                  & $24.44 \pm 3.85$                  & $73.78 \pm 1.27$                  & $41.00 \pm 3.17$ \\
 & MAD              & $72.20 \pm 5.74$                  & $25.56 \pm 1.92$                  & $20.00 \pm 3.33$                  & $79.00 \pm 0.70$                  & $38.84 \pm 1.46$ \\
 & SoM              & $79.20 \pm 0.20$                  & $28.89 \pm 4.95$                  & \underline{$26.67 \pm 3.33$}       & $77.93 \pm 4.42$                  & \underline{$44.57 \pm 1.61$} \\
 & DMAD             & \underline{$80.47 \pm 0.76$}       & $28.89 \pm 1.92$                  & $24.44 \pm 5.09$                  & $78.11 \pm 0.70$                  & $43.90 \pm 1.45$ \\
 & \textbf{DynaDebate (Ours)} & \textbf{82.17 $\pm$ 1.42}   & \textbf{35.56 $\pm$ 5.09}          & \textbf{27.78 $\pm$ 8.39}          & \textbf{82.44 $\pm$ 0.51}          & \textbf{46.38 $\pm$ 1.01} \\
\midrule

\multirow{8}{*}{\textbf{GPT-4o-mini}}
 & CoT              & $69.60 \pm 0.85$                  & $8.89 \pm 1.92$                   & $8.89 \pm 1.92$                   & $82.00 \pm 0.67$                  & $42.13 \pm 0.84$ \\
 & CoT-SC           & $68.73 \pm 4.97$                  & $11.11 \pm 1.92$                  & $11.11 \pm 1.92$                  & \underline{$83.00 \pm 1.20$}       & \textbf{47.99 $\pm$ 0.18} \\
 & Self-Refine      & $69.93 \pm 0.90$                  & $8.89 \pm 1.92$                   & $10.00 \pm 0.00$                  & $75.22 \pm 1.68$                  & $40.70 \pm 2.41$ \\
 & Self-Contrast    & $71.27 \pm 1.30$                  & \underline{$12.22 \pm 5.09$}       & \textbf{16.67 $\pm$ 3.33}          & $61.22 \pm 1.92$                  & \underline{$44.50 \pm 2.28$} \\
 & MAD              & $64.40 \pm 5.40$                  & $10.00 \pm 3.33$                  & $8.89 \pm 3.85$                   & $81.33 \pm 0.88$                  & $39.44 \pm 1.95$ \\
 & SoM              & \underline{$72.27 \pm 2.20$}       & $8.89 \pm 6.95$                   & \underline{$15.55 \pm 3.85$}       & $80.20 \pm 1.06$                  & $41.07 \pm 1.56$ \\
 & DMAD             & $70.47 \pm 1.30$                  & $8.89 \pm 5.09$                   & $12.22 \pm 1.92$                  & $81.22 \pm 2.17$                  & $43.60 \pm 2.66$ \\
 & \textbf{DynaDebate (Ours)} & \textbf{73.40 $\pm$ 0.20}   & \textbf{14.44 $\pm$ 5.09}          & $13.33 \pm 0.00$                  & \textbf{84.44 $\pm$ 1.02}          & \textbf{47.99 $\pm$ 0.14} \\
\midrule

\multirow{8}{*}{\textbf{Llama-3.3-70B-Instruct}}
 & CoT              & $67.80 \pm 0.53$                  &  \underline{24.44 $\pm$ 3.85}                  & \underline{$4.44 \pm 1.93$}        & $86.62 \pm 0.54$                  & $60.64 \pm 0.46$ \\
 & CoT-SC           & $60.27 \pm 1.50$                  & \textbf{25.56 $\pm$ 1.93}          & \underline{$4.44 \pm 1.93$}        & $86.51 \pm 1.19$                  & $63.05 \pm 0.56$ \\
 & Self-Refine      & $69.20 \pm 0.40$                  & $23.33 \pm 3.34$                  & \textbf{6.67 $\pm$ 0.00}           & $86.67 \pm 0.58$                  & $60.94 \pm 0.36$ \\
 & Self-Contrast    & $70.27 \pm 0.83$                  &  \underline{24.44 $\pm$ 1.57}                  & \textbf{6.67 $\pm$ 0.00}           & \textbf{87.78 $\pm$ 0.63}          & $63.62 \pm 1.00$ \\
 & MAD              & $53.20 \pm 2.78$                  & $22.22 \pm 1.92$                  & \underline{$4.44 \pm 1.93$}        & $85.67 \pm 1.15$                  & $60.94 \pm 0.49$ \\
 & SoM              & $69.53 \pm 0.50$                  & $22.22 \pm 1.92$                  & \textbf{6.67 $\pm$ 3.34}           & $85.78 \pm 0.39$                  & \underline{$64.81 \pm 0.28$} \\
 & DMAD             & \underline{$72.00 \pm 0.40$}       & \underline{24.44 $\pm$ 1.93}                  & \textbf{6.67 $\pm$ 0.00}           & $86.44 \pm 1.39$                  & $62.76 \pm 0.79$ \\
 & \textbf{DynaDebate (Ours)} & \textbf{73.27 $\pm$ 0.31}   & \textbf{25.56 $\pm$ 3.85}       & \textbf{6.67 $\pm$ 0.00}           & \underline{$86.96 \pm 0.08$}       & \textbf{66.27 $\pm$ 0.30} \\
\bottomrule
\end{tabular}
}
\caption{
Main results of DynaDebate compared with baselines across three backbone models and five benchmarks. Results are mean $\pm$ std (\%) over three runs.
\textbf{Bold} indicates the best result per column per model block; \underline{underline} denotes the second-best.
}
\label{tab:main_results}
\vspace{-4pt}
\end{table*}

\subsection{Results}

DynaDebate attains the best or tied-best result on 13 of the 15 benchmark--backbone combinations, achieving superior or highly competitive performance across the majority of benchmarks. More revealing than the win count is a consistent pattern: its margin over baselines widens as problems demand a narrower set of valid solution strategies. The lead is slim on MMLU, where most reasoning paths converge on the same answer, but substantial on AIME across all three backbones, where a single misjudged entry point dooms an entire trajectory. This is precisely what the homogeneity hypothesis predicts---when the space of plausible-but-wrong strategies is large, a debate seeded with diverse paths is far more likely to contain a correct line of attack---indicating that the gains stem from diversification, not incidental prompting effects.

Crucially, the effect is not confined to mathematics. On GPQA, which requires graduate-level scientific reasoning, DynaDebate attains the highest accuracy on all three backbones, and it leads on MMLU for two of three. That the same mechanism helps both symbolic derivation and knowledge-intensive expert reasoning indicates that premature convergence is a domain-general failure mode of multi-agent debate, not an artifact of arithmetic. The baselines further localize the source of the gains: standard MAD methods (MAD, SoM) are weakest on exactly the hard mathematical tasks where shared initialization bias is most dangerous, a predictable consequence of majority voting, which amplifies rather than corrects an error that all agents inherit from a common starting point. DynaDebate removes this correlated bias upstream by allocating genuinely distinct reasoning paths before debate begins.

Two exceptions are instructive rather than anomalous. Self-Contrast edges out DynaDebate on Llama-3.3-70B MMLU (87.78\% vs.\ 86.96\%), where a capable backbone on a coverage-oriented task gains little from structured debate, and on GPT-4o-mini AIME 2025 (13.33\% vs.\ 16.67\%); both fall outside the difficulty-and-diversity regime our hypothesis targets. Finally, Appendix~\ref{app:model_scale} shows that on reasoning-intensive tasks a Qwen3-8B model with DynaDebate matches or exceeds a Qwen3-32B model running standard CoT (e.g., $+13.3$ points on AIME 2025), suggesting that disciplined diversification can partially substitute for raw model scale.


\subsection{Ablation Studies}
\label{sec:ablation}

To validate the contribution of each core mechanism in \textbf{DynaDebate}, we conducted ablation studies using Qwen3-8B as the base model across multiple independent trials, reporting mean $\pm$ standard deviation to ensure statistical reliability. Note that the Biography dataset is excluded from this analysis, as for this task, we did not employ the full DynaDebate framework, but instead utilized only the Path Generation and Allocation module (see Appendix~\ref{app:datasets} for details). We define three variants by removing specific components:

\begin{itemize}
    \item \textbf{w/o Path Generation and Allocation}: Removes the initialization stage; agents enter debate without guided reasoning paths.
    \item \textbf{w/o Process-Centric Debate}: Removes the step-by-step peer review mechanism.
    \item \textbf{w/o Verification}: Removes the trigger-based verification agent.
\end{itemize}

\begin{table*}[t]
\centering
\resizebox{\textwidth}{!}{%
\begin{tabular}{l|ccccc}
\toprule
\textbf{Method} & \textbf{AIME 2024} & \textbf{AIME 2025} & \textbf{MATH500} & \textbf{MMLU} & \textbf{GPQA} \\
\midrule
w/o Path Generation and Allocation & $28.89 \pm 1.92$ & $21.11 \pm 3.85$ & $80.73 \pm 0.61$ & $81.22 \pm 0.51$ & $46.01 \pm 0.65$ \\
w/o Process-Centric Debate         & $33.33 \pm 3.34$ & $25.55 \pm 3.85$ & $79.53 \pm 0.23$ & $82.00 \pm 0.67$ & $44.41 \pm 1.13$ \\
w/o Verification                   & $28.89 \pm 1.92$ & $24.44 \pm 1.93$ & $80.93 \pm 0.90$ & $80.89 \pm 0.84$ & $45.04 \pm 0.26$ \\
\midrule
\textbf{DynaDebate (Full)} & \textbf{35.56 $\pm$ 5.09} & \textbf{27.78 $\pm$ 8.39} & \textbf{82.17 $\pm$ 1.42} & \textbf{82.44 $\pm$ 0.51} & \textbf{46.38 $\pm$ 1.01} \\
\bottomrule
\end{tabular}
}
\caption{Ablation study on Qwen3-8B. We report mean $\pm$ std accuracy (\%) across five benchmarks over multiple independent runs.}
\label{tab:ablation}
\vspace{-4pt}
\end{table*}

Table~\ref{tab:ablation} summarizes the results.

\paragraph{Impact of Path Generation and Allocation.}
Removing Path Generation and Allocation produces a sharply difficulty-dependent effect: accuracy falls by over six points on AIME 2024 ($35.56 \to 28.89$) and AIME 2025 ($27.78 \to 21.11$), but only marginally on MATH500 ($-1.44$) and MMLU ($-1.22$). Diverse initialization matters most exactly when a task offers non-obvious entry points that agents would otherwise miss, confirming that the module prevents premature convergence on the most salient-but-wrong trajectory rather than adding a uniform boost. Appendix~\ref{app:trigger_stats} corroborates this: path coverage rate exceeds single-path rate by 7--20 points, so the allocated paths are functionally complementary rather than redundant paraphrases.

\paragraph{Impact of Process-Centric Debate.}
Process-Centric Debate behaves oppositely: its removal degrades every benchmark (MATH500 $-2.64$, AIME 2024 $-2.23$, AIME 2025 $-2.23$, GPQA $-1.97$). This domain-agnostic cost shows that step-level auditing repairs a failure invisible to the outcome-level comparison used by prior MAD methods---intermediate errors that survive when only final answers are contrasted---making it a general improvement to debate quality rather than a task-specific heuristic.

\paragraph{Impact of Verification.}
Verification, in contrast, is a targeted safety net: its removal costs the most on the hardest tasks (AIME 2024 $-6.67$, AIME 2025 $-3.34$) and at most $1.55$ points elsewhere. This tracks the trigger frequencies in Appendix~\ref{app:trigger_stats}, where the module fires on roughly 80--90\% of AIME problems---the highest rate of any benchmark---because external tools are invoked precisely when agents deadlock on difficult problems. Verification thus contributes where it is genuinely needed rather than uniformly.

\subsection{Discussions }
\label{sec:analysis}

\subsubsection{Discussions on Diversity Analysis}

\begin{figure*}[t]
    \centering
    \begin{subfigure}[t]{0.495\textwidth}
        \centering
        \includegraphics[width=\linewidth]{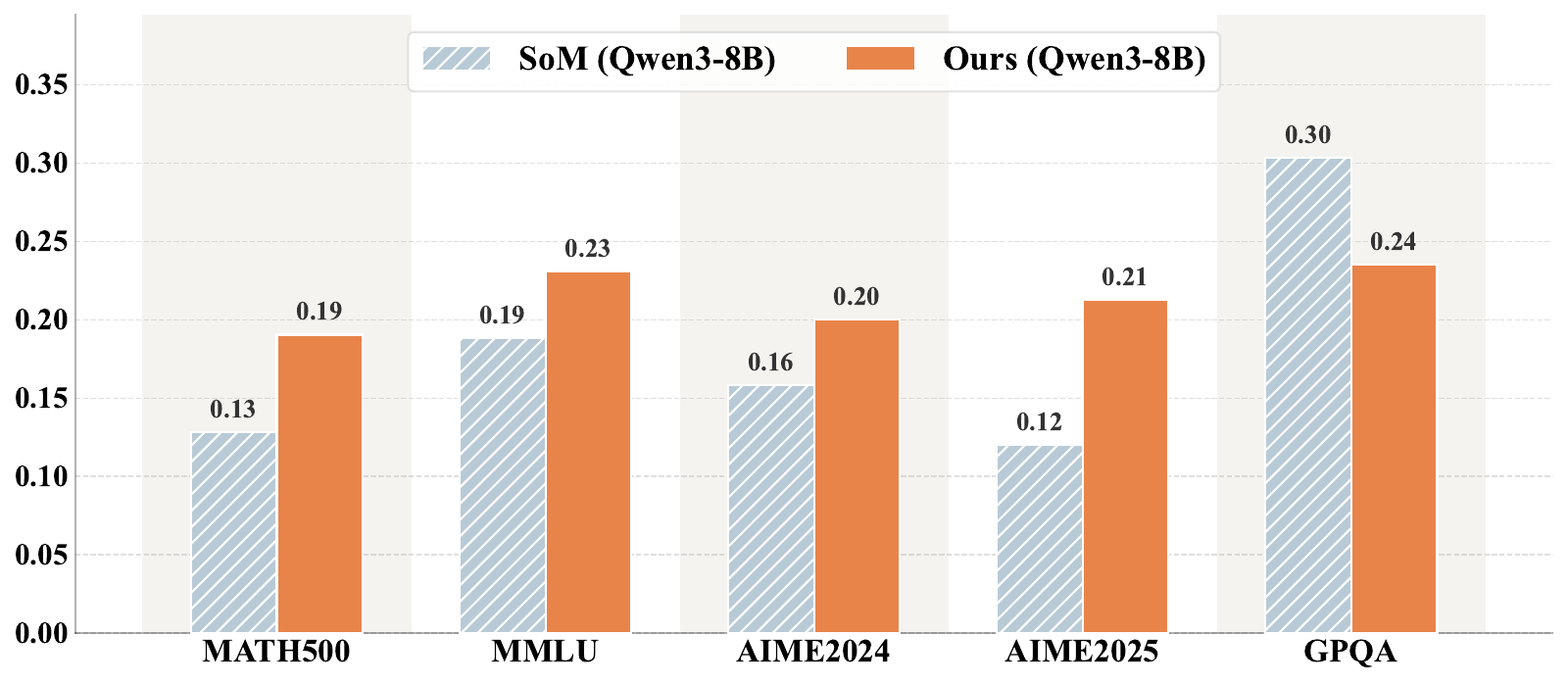}
        \caption{Intra-diversity across datasets and backbone models.}
        \label{fig:intra_diversity}
    \end{subfigure}
    \hfill
    \begin{subfigure}[t]{0.495\textwidth}
        \centering
        \includegraphics[width=\linewidth]{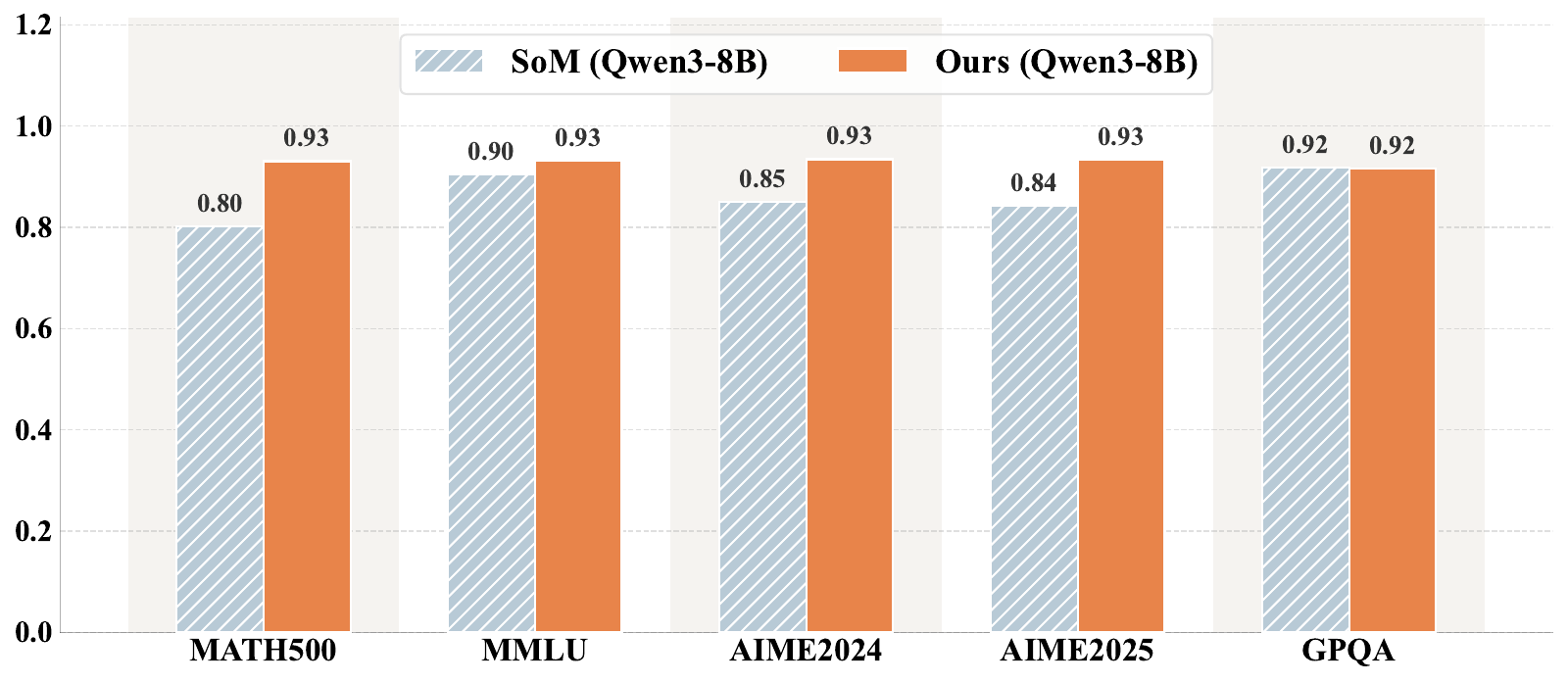}
        \caption{Structural non-overlap across datasets and backbone models.}
        \label{fig:structural_nonoverlap}
    \end{subfigure}

    \caption{
    Diversity analysis of multi-agent reasoning.
    (a) Intra-diversity measures semantic variation among agents.
    (b) Structural non-overlap measures diversity at the reasoning-structure level.
    Our method consistently improves both metrics across datasets, indicating structured and effective diversification.
    }
    \label{fig:diversity_analysis}
    \vspace{-4pt}
\end{figure*}

To quantify the diversity of multi-agent reasoning, we consider two complementary metrics: \textbf{Intra-diversity} and \textbf{Structural Non-overlap}, together with task accuracy to assess whether diversity is beneficial rather than noisy.

\paragraph{Intra-diversity.}
Intra-diversity measures semantic variation among agents' reasoning for the same problem. Given the final responses $\{r_1, \dots, r_N\}$ from $N$ agents, we compute the average pairwise cosine distance between their TF--IDF representations:

\begin{equation}
\begin{aligned}
\mathrm{IntraDiv}(x)
&= \frac{2}{N(N-1)} \sum_{i<j} \big(1 \\
&\quad - \cos\big(\phi(r_i), \phi(r_j)\big)\big),
\end{aligned}
\end{equation}
where $\phi(\cdot)$ denotes TF--IDF encoding. Higher values indicate that agents explore more diverse semantic reasoning paths instead of converging to similar explanations.

\paragraph{Structural Non-overlap.}
While semantic diversity captures surface-level variation, it does not necessarily reflect differences in reasoning structure. To address this, we further measure Structural Non-overlap. For each response $r_i$, we segment the reasoning into step-level units $S_i$, and compute their Jaccard overlap:
\begin{equation}
\mathrm{NonOverlap}(x) = 1 - \frac{2}{N(N-1)} \sum_{i<j} \frac{|S_i \cap S_j|}{|S_i \cup S_j|}.
\end{equation}
Higher values indicate that agents rely on distinct reasoning structures rather than paraphrasing similar logic.

As shown in Figure~\ref{fig:intra_diversity}, our method achieves higher intra-diversity across most benchmarks, including MATH500, MMLU, AIME 2024, and AIME 2025. On GPQA, SoM exhibits slightly higher intra-diversity (0.30 vs.\ 0.24), potentially because surface-level response variation on expert-knowledge problems is less constrained by structured path guidance. Overall, these results indicate that the proposed mechanism effectively encourages agents to explore diverse reasoning paths rather than respond homogeneously.

Figure~\ref{fig:structural_nonoverlap} further demonstrates that step non-overlap is consistently improved across most benchmarks, indicating that the increased diversity arises from genuinely different reasoning structures rather than superficial variations. On GPQA, both methods reach the same step non-overlap score (0.92), suggesting that structural diversity gains are most pronounced on tasks requiring extended multi-step derivation. Importantly, these improvements do not come at the expense of task performance, as the overall accuracy remains competitive and, in some cases, even improves.

Together, these results demonstrate that our approach promotes structured and effective diversity in multi-agent reasoning.
To provide concrete insights into how agents correct errors and diversify reasoning paths in practice, we present detailed qualitative case studies in Appendix~\ref{app:case_study}.

\subsubsection{Discussions on Effect of the Number of Agents.}
\begin{figure}[t]
    \vspace{-2pt}
    \centering
    \includegraphics[width=\linewidth]{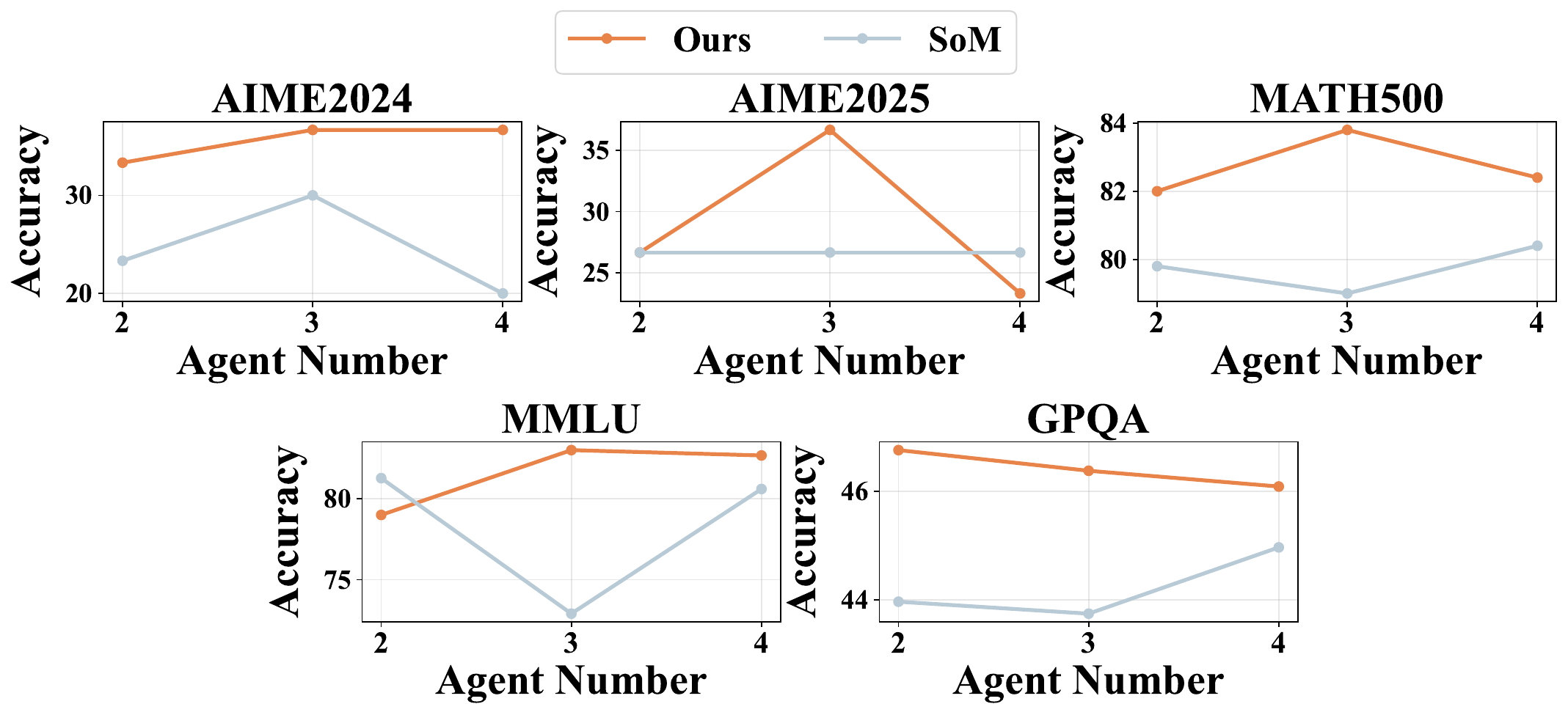}
    \caption{Performance comparison under different numbers of agents.}
    \label{fig:agent_num_ablation}
\end{figure}

Figure~\ref{fig:agent_num_ablation} shows a non-monotonic trend: moving from two to three agents yields consistent gains on most benchmarks, while GPQA peaks at two agents, suggesting the model can generate only two high-quality independent paths for this domain. Beyond three agents, performance plateaus or reverses: once agents outnumber valid paths $K$, the round-robin allocation assigns duplicate initializations, and the model cannot reliably produce additional independent alternatives. Three agents thus represents the optimal configuration.

\subsubsection{Discussions on Effect of Debate Rounds.}
\begin{figure}[t]
    \vspace{-2pt}
    \centering
    \includegraphics[width=\linewidth]{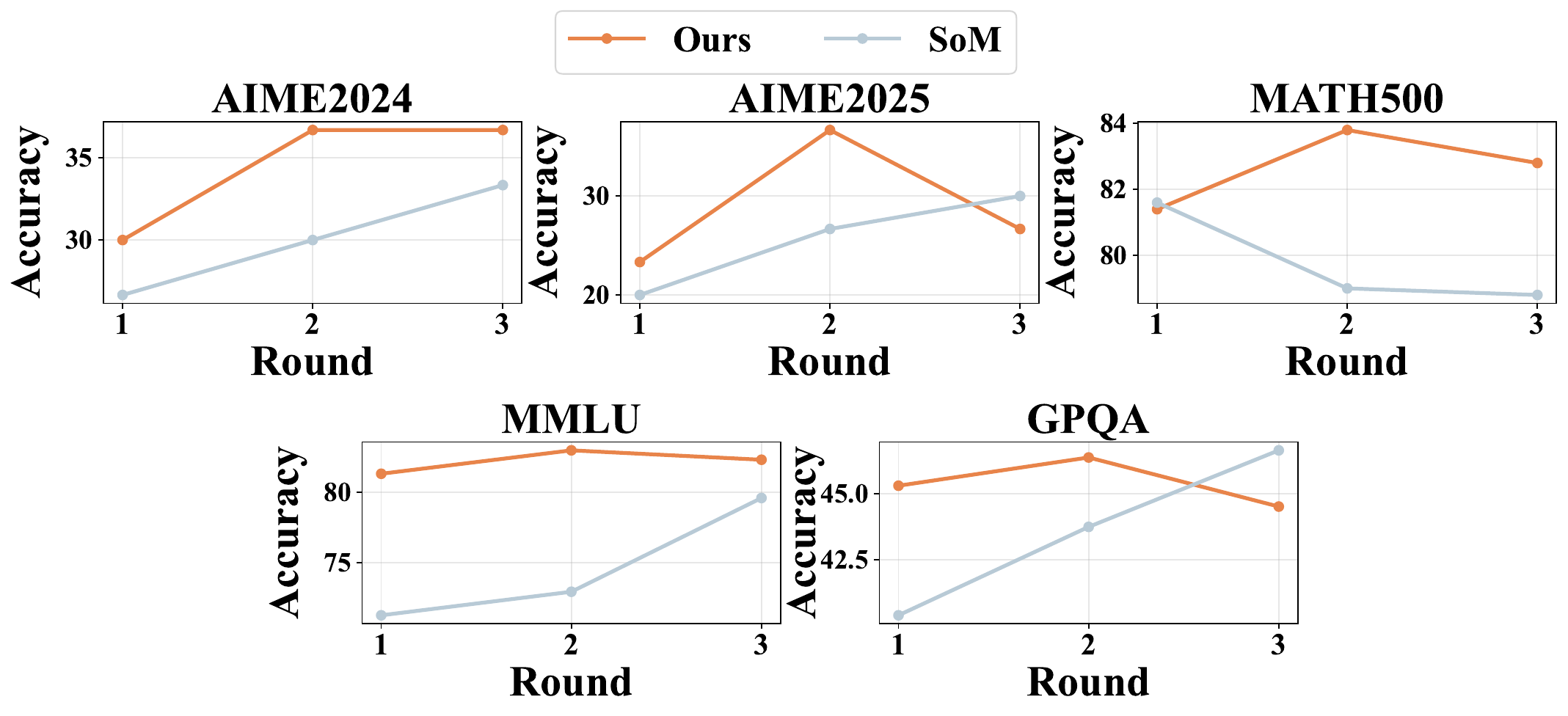}
    \caption{Effect of the number of debate rounds $N$ on performance across benchmarks.}
    \label{fig:round_ablation}
\end{figure}

Figure~\ref{fig:round_ablation} shows that hard mathematical tasks peak at Round 2 and degrade at Round 3, as growing context length impairs coherent reasoning and prolonged peer exposure reduces the independence of agent contributions. Knowledge-intensive tasks are more robust, as factual correctness stabilizes once established. Crucially, DynaDebate reaches peak performance by Round 2 on most benchmarks, while SoM requires Round 3 to achieve comparable results—and still falls short—demonstrating that structured initialization enables faster and more efficient convergence.

\subsubsection{Discussions on Computation Cost vs. Performance.}

\begin{figure}[tb]
    \centering
    \includegraphics[width=\linewidth]{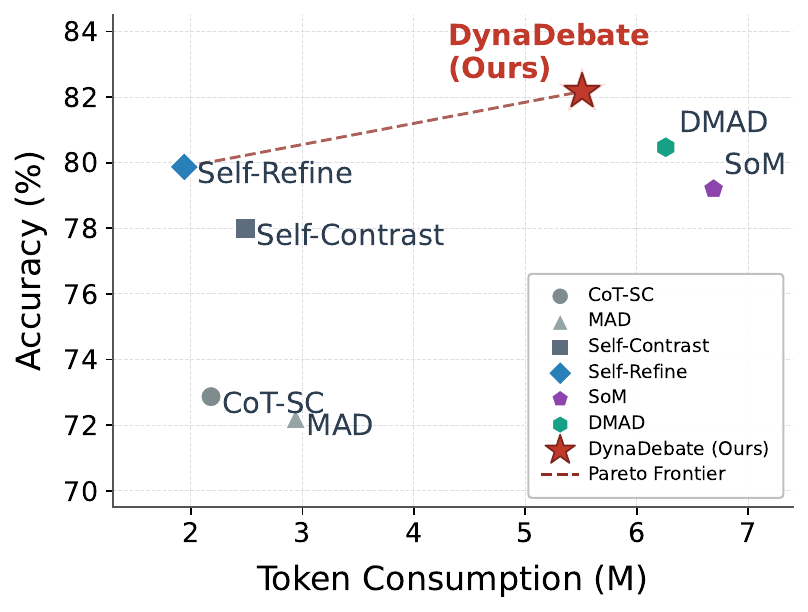}
    \caption{Accuracy--token Pareto frontier on MATH500. DynaDebate consistently achieves higher accuracy at comparable token budgets relative to baselines.}
    \label{fig:pareto}
\end{figure}

As shown in Figure~\ref{fig:pareto}, DynaDebate consistently occupies a favorable position on the accuracy--token Pareto frontier, achieving higher accuracy at comparable token budgets relative to all baselines. Moreover, this overhead is selective rather than uniform, as verification triggers predominantly on the hardest problems (Appendix~\ref{app:trigger_stats}). Per-round token costs appear in Appendix~\ref{app:cost}.

\vspace{-4pt}
\section{Conclusion}
\label{sec:conclusion}
We proposed \textbf{DynaDebate} to address reasoning homogeneity in multi-agent debate. By integrating dynamic path generation and allocation, process-centric debate, and trigger-based verification, our framework achieves superior or highly competitive performance across five benchmarks. We hope this work encourages future exploration of structured initialization heterogeneity and process-level verification in collaborative LLM reasoning systems.

\section{Limitations}
A primary limitation of our work lies in the trade-off between computational efficiency and performance gains on tasks with lower reasoning complexity. On near-saturated benchmarks (e.g., GSM8K; see Appendix~\ref{app:gsm8k}), where current models already exhibit high proficiency, the marginal performance improvements provided by DynaDebate are less significant. Since errors in these tasks rarely stem from insufficient reasoning capability, the elaborate mechanisms of path allocation and multi-round debate may not yield substantial benefits. Consequently, applying the full multi-agent framework to such simpler problems introduces additional token consumption without a proportional return in accuracy compared to standard prompting methods.

\bibliography{main}
\clearpage

\appendix
\section{Pseudocode}
\label{app:pseudocode}

Algorithm~\ref{alg:dynadebate} outlines the core execution flow of the DynaDebate framework, including Path Generation, Process-Centric Debate, and Verification triggering mechanisms.

\begin{algorithm}[t]
\caption{DynaDebate Framework for Multi-Agent Reasoning}
\label{alg:dynadebate}
\begin{algorithmic}[1]
\Require Input query $q$, Agents $\mathcal{A} = \{a_1, \dots, a_N\}$, Max Rounds $T$
\Ensure Final Answer $y$

\State $\mathcal{P} \leftarrow \Phi_{\text{gen}}(q)$; \quad $K \leftarrow |\mathcal{P}|$
\For{$i = 1, \dots, N$}
    \State $\text{Path}(a_i) \leftarrow p_{(i-1 \pmod K) + 1}$
\EndFor

\State $H_0 \leftarrow \emptyset$
\For{$t = 1, \dots, T$}
    \For{$i = 1, \dots, N$}
        \If{$t = 1$}
            \State $r_{i,t} \leftarrow a_i(q \mid \text{Path}(a_i))$
        \Else
            \State $r_{i,t} \leftarrow a_i(q, H_{t-1})$
        \EndIf
    \EndFor
    \State $H_t \leftarrow H_{t-1} \cup \{r_{i,t}\}_{i=1}^N$

    \If{$\text{Trigger}(H_t)$}
        \State $o_t \leftarrow \Phi_{\text{ver}}(q, H_t)$
        \State $H_t \leftarrow H_t \cup \{o_t\}$
    \EndIf

\EndFor
\State \Return $\text{ExtractAnswer}(H_t)$
\end{algorithmic}
\end{algorithm}

\section{GSM8K Results}
\label{app:gsm8k}

We further evaluate DynaDebate on the widely-used arithmetic reasoning benchmark GSM8K~\cite{cobbe2021gsm8k}.
As shown in Table~\ref{tab:gsm8k_results}, all evaluated methods achieve above 94\% accuracy, indicating that GSM8K is approaching saturation for models at this scale.
In this regime, the collaborative overhead of multi-agent debate yields only marginal incremental benefit, which is consistent with our hypothesis that DynaDebate's performance gains scale with task difficulty.

\begin{table}[h]
\centering
\resizebox{\columnwidth}{!}{%
\begin{tabular}{llc}
\toprule
\textbf{Model} & \textbf{Method} & \textbf{GSM8K} \\
\midrule
\multirow{8}{*}{\textbf{Qwen3-8B}}
 & CoT              & $95.11 \pm 0.96$ \\
 & CoT-SC           & \underline{$95.33 \pm 0.58$} \\
 & Self-Refine      & $92.78 \pm 0.69$ \\
 & Self-Contrast    & $94.11 \pm 0.84$ \\
 & MAD              & $84.33 \pm 5.77$ \\
 & SoM              & \textbf{95.56 $\pm$ 0.51} \\
 & DMAD             & $94.44 \pm 0.51$ \\
 & \textbf{DynaDebate (Ours)} & $95.00 \pm 0.67$ \\
\midrule
\multirow{8}{*}{\textbf{GPT-4o-mini}}
 & CoT              & $93.33 \pm 0.58$ \\
 & CoT-SC           & $93.78 \pm 0.84$ \\
 & Self-Refine      & $89.89 \pm 0.19$ \\
 & Self-Contrast    & $90.00 \pm 3.17$ \\
 & MAD              & $87.00 \pm 1.20$ \\
 & SoM              & \underline{$94.44 \pm 0.84$} \\
 & DMAD             & $93.67 \pm 0.67$ \\
 & \textbf{DynaDebate (Ours)} & \textbf{95.00 $\pm$ 0.67} \\
\midrule
\multirow{8}{*}{\textbf{Llama-3.3-70B-Instruct}}
 & CoT              & \underline{$98.33 \pm 0.58$} \\
 & CoT-SC           & $98.00 \pm 1.00$ \\
 & Self-Refine      & $95.67 \pm 0.58$ \\
 & Self-Contrast    & $96.56 \pm 0.42$ \\
 & MAD              & $90.67 \pm 3.21$ \\
 & SoM              & \underline{$98.33 \pm 0.58$} \\
 & DMAD             & \textbf{98.67 $\pm$ 0.58} \\
 & \textbf{DynaDebate (Ours)} & \underline{$98.33 \pm 0.58$} \\
\bottomrule
\end{tabular}
}
\caption{GSM8K results across three backbone models. Results are mean $\pm$ std (\%) over three runs. \textbf{Bold} indicates the best result per model block; \underline{underline} denotes the second-best.}
\label{tab:gsm8k_results}
\end{table}

\section{Dataset Details}
\label{app:datasets}

We evaluate our proposed framework on seven datasets spanning mathematical reasoning, general knowledge, expert knowledge reasoning, and factual correctness.

\subsection{Mathematical Reasoning Tasks}
\begin{itemize}
    \item \textbf{GSM8K}~ \cite{cobbe2021gsm8k}: A benchmark consisting of high-quality, linguistically diverse grade school math word problems. We randomly sampled 300 questions from the test set for evaluation. This dataset serves as a baseline to assess the model's capability in basic multi-step reasoning.
    \item \textbf{MATH500}~ \cite{lightman2023lets}: A rigorous subset of the MATH dataset, containing 500 challenging problems uniformly sampled across seven subjects. Unlike GSM8K, MATH500 requires complex problem-solving skills and domain-specific knowledge.
    \item \textbf{AIME 2024-2025}~ \cite{aime2024,aime2025}: The American Invitational Mathematics Examination (AIME) represents the pinnacle of high-school mathematical competitions. We utilize problems from the 2024 and 2025 exams to test the model's performance on extremely hard, out-of-distribution reasoning tasks.
\end{itemize}

\subsection{General Knowledge Tasks}
\begin{itemize}
    \item \textbf{MMLU}~ \cite{hendrycks2020measuring}: The Massive Multitask Language Understanding benchmark covers 57 subjects across STEM, the humanities, and social sciences. We randomly sampled 300 questions to assess the breadth of the model's world knowledge.
\end{itemize}

\subsection{Expert Knowledge Reasoning Tasks}
\begin{itemize}
    \item \textbf{GPQA}~\cite{rein2023gpqa}: The Graduate-Level Google-Proof Q\&A benchmark is a challenging multiple-choice dataset designed to evaluate expert-level scientific knowledge and reasoning. Questions are authored by domain experts in biology, physics, and chemistry, and are specifically designed to be difficult even for those with internet access (hence ``Google-proof''). We use the GPQA Main subset, which contains 448 questions spanning biology, physics, and chemistry. Standard accuracy is used as the evaluation metric.
\end{itemize}

\subsection{Factual Correctness Tasks}
\begin{itemize}
    \item \textbf{Biography}~ \cite{du2023improving}: We randomly selected 100 figures to evaluate hallucination mitigation. The evaluation protocol follows the methodology proposed in the SoM paper.
\paragraph{Special Evaluation details}
    For each figure, an independent LLM acts as a \textbf{judge} to compare the generated biography against reference facts claim-by-claim. For each fact, the judge determines whether the generated biography is consistent with the fact and assigns one of three labels:
    \begin{itemize}
        \item \textbf{yes}: The fact is explicitly stated or clearly implied.
        \item \textbf{no}: The fact is missing or contradicts the biography.
        \item \textbf{uncertain}: The biography is too vague to determine consistency.
    \end{itemize}

    The metric calculates accuracy based only on facts labeled as ``yes'' or ``no'', ignoring ``uncertain'' cases:
    \begin{equation}
        \text{Accuracy}_{\text{Yes/No}} = \frac{N_{\text{yes}}}{N_{\text{yes}} + N_{\text{no}}}
    \end{equation}
    This metric reflects the model's accuracy when it provides explicit factual statements.
\paragraph{Component-Based Agent Allocation for Biography Generation}

For the Biography benchmark, we adopt a component-based agent allocation strategy,
denoted as allocate\_component, to structure multi-agent generation.
Instead of asking agents to generate a full biography independently, we first decompose
a biography into a small number of semantically distinct components (depends on number of agents),
such as core research focus, major contributions, and professional roles.

Each agent is assigned exactly one component and is instructed to generate content
only for that component.
Generation proceeds in multiple rounds: agents first produce initial component drafts,
and then refine their own components by referencing others for context,
while being prohibited from introducing new facts outside their responsibilities.

After all rounds, the generated components are merged into a single biography through
a lightweight integration step that removes duplication and ensures coherence,
without adding any new information.
\end{itemize}

\section{Biography Benchmark Results}
\label{app:biography}

\paragraph{Benchmark Description.}
The Biography benchmark, introduced in SoM~\cite{du2023improving}, evaluates hallucination mitigation in open-ended text generation using Fact-level Accuracy. We randomly selected 100 public figures and evaluate biographies generated by each method. For this task, DynaDebate employs only the Path Generation and Allocation module using a component-based strategy (see Appendix~\ref{app:datasets} for details), as the open-ended generation nature of this task does not require the full debate pipeline.

\paragraph{Results.}
Table~\ref{tab:biography_results} presents the full experimental results across all three backbone models. Results are reported as mean $\pm$ standard deviation (\%) over three independent runs. Self-Contrast has no Biography results as it is not designed for open-ended generation tasks.

\begin{table}[h]
\centering
\resizebox{\columnwidth}{!}{%
\begin{tabular}{llc}
\toprule
\textbf{Model} & \textbf{Method} & \textbf{Biography} \\
\midrule
\multirow{8}{*}{\textbf{Qwen3-8B}}
 & CoT              & $53.19 \pm 4.83$ \\
 & CoT-SC           & $44.89 \pm 0.90$ \\
 & Self-Refine      & \textbf{58.81 $\pm$ 1.27} \\
 & MAD              & $46.30 \pm 0.96$ \\
 & SoM              & $42.81 \pm 0.98$ \\
 & DMAD             & $38.98 \pm 0.92$ \\
 & \textbf{DynaDebate (Ours)} & \underline{$53.29 \pm 2.33$} \\
\midrule
\multirow{8}{*}{\textbf{GPT-4o-mini}}
 & CoT              & $86.87 \pm 0.66$ \\
 & CoT-SC           & $86.51 \pm 1.17$ \\
 & Self-Refine      & \textbf{92.81 $\pm$ 0.92} \\
 & MAD              & $86.03 \pm 0.25$ \\
 & SoM              & $76.58 \pm 0.96$ \\
 & DMAD             & $79.58 \pm 1.21$ \\
 & \textbf{DynaDebate (Ours)} & \underline{$90.79 \pm 1.62$} \\
\midrule
\multirow{8}{*}{\textbf{Llama-3.3-70B-Instruct}}
 & CoT              & $87.12 \pm 0.34$ \\
 & CoT-SC           & $87.88 \pm 0.22$ \\
 & Self-Refine      & \underline{91.33 $\pm$ 0.36} \\
 & MAD              & $84.70 \pm 0.50$ \\
 & SoM              & $62.89 \pm 1.52$ \\
 & DMAD             & $79.20 \pm 0.26$ \\
 & \textbf{DynaDebate (Ours)} & \textbf{93.57 $\pm$ 0.54} \\
\bottomrule
\end{tabular}
}
\caption{Biography benchmark results (Fact-level Accuracy, \%). \textbf{Bold} indicates the best performance per model block; \underline{underline} denotes the second-best. ``--'' indicates the method is not applicable to open-ended generation.}
\label{tab:biography_results}
\end{table}

\paragraph{Analysis.}
The results on the Biography benchmark reveal an interesting pattern that varies across backbone models.
For Qwen3-8B and GPT-4o-mini, Self-Refine attains the highest score ($58.81\%$ and $92.81\%$, respectively), while DynaDebate achieves the second-best performance on both models ($53.29\%$ and $90.79\%$).
On Llama-3.3-70B-Instruct, DynaDebate achieves the best performance ($93.57\%$), with Self-Refine as the second-best ($91.33\%$).
The strong performance of Self-Refine on smaller models is consistent with its design, which focuses on iterative output polishing and is well-suited for open-ended generation tasks that emphasize factual completeness.
Notably, SoM performs considerably lower than other methods on Llama-3.3-70B-Instruct ($62.89\%$), suggesting that the Society of Mind approach may not transfer well to this backbone for factual generation tasks.

\section{Baseline Implementations}
\label{app:baselines}

To demonstrate the effectiveness of DynaDebate, we compare it against a comprehensive set of baselines.
\vspace{-4pt}
\subsection{Single-Agent Strategies}
\begin{itemize}
    \item \textbf{Chain-of-Thought (CoT)}~\cite{wei2022chain}: The standard prompting approach where the model is instructed to ``think step by step'' before generating the final answer.
    \item \textbf{CoT-SC (Self-Consistency)}~\cite{wang2022self}: An ensemble strategy that samples $k$ distinct reasoning paths and determines the final answer by a majority vote.
    \item \textbf{Self-Refine}~\cite{madaan2023self}: An iterative feedback mechanism where a single agent generates an initial output, critiques its own reasoning, and then refines the answer based on self-generated feedback.
\end{itemize}

\subsection{Multi-Agent Frameworks}
\begin{itemize}
    \item \textbf{MAD (Multi-Agent Debate)}~\cite{liang2023encouraging}: This method is implemented by inducing divergent results through participants with different personas: an Affirmative participant (``Angel'') and a Negative participant (``Devil'') provide answers to a Judge participant. The Judge is responsible for managing the process and deriving the final solution. The Judge possesses additional autonomy and can terminate the debate early if satisfied with the answer.
    \item \textbf{SoM (Society of Mind)}~\cite{du2023improving}: This framework proposes a multi-agent decision (MAD) approach where multiple agents achieve efficient collaboration by passing answers to one another. The method allows for the summarization of answers before adding them to the history record, enabling agents to call upon this condensed information in subsequent rounds.
    \item \textbf{DMAD}~\cite{liu2025breaking}: \textbf{Diversified Multi-Agent Debate (DMAD)} is a novel and robust framework designed to guide agents in employing diverse reasoning methods. It aims to break thinking stereotypes and patterns, thereby enhancing overall reasoning performance.
\end{itemize}

\section{Impact of Model Scale: Can Small Models Outperform Large Models?}
\label{app:model_scale}

\begin{table*}[h]
    \centering
    \small
    \begin{tabular}{lcccccc}
        \toprule
        \textbf{Model \& Method} & \textbf{AIME24} & \textbf{AIME25} & \textbf{MATH500} & \textbf{GSM8K} & \textbf{MMLU} & \textbf{Bio} \\
        \midrule
        Qwen3-32B (CoT) & 36.67 & 23.33 & 80.80 & \textbf{95.67} & \textbf{83.33} & \textbf{59.09} \\
        \textbf{Qwen3-8B (Ours)} & \textbf{36.67} & \textbf{36.67} & \textbf{83.80} & \textbf{95.67} & 83.00 & 52.72 \\
        \midrule
        \textit{Gap (8B vs 32B)} & \textit{0.00} & \textit{+13.34} & \textit{+3.00} & \textit{0.00} & \textit{-0.33} & \textit{-6.37} \\
        \bottomrule
    \end{tabular}
     \caption{Performance comparison between Qwen3-8B (with DynaDebate) and Qwen3-32B (with Standard CoT). \textbf{Bold} indicates the superior performance.}
    \label{tab:scale_comparison}
\end{table*}

A critical question in multi-agent research is whether structured collaboration can compensate for a lack of intrinsic parameter scale. To investigate this, we compare the performance of a smaller model using our framework against a significantly larger model using standard reasoning methods.

We conduct a comparative experiment using the Qwen3 family to control for training data and architecture differences:
\begin{itemize}
    \item \textbf{Small Model (Ours):} \textbf{Qwen3-8B} equipped with the \textsc{DynaDebate} framework.
    \item \textbf{Large Model (Baseline):} \textbf{Qwen3-32B} utilizing standard Chain-of-Thought (CoT) prompting.
\end{itemize}
Both models were evaluated across the same diverse set of benchmarks spanning mathematical reasoning, general knowledge, and hallucination mitigation.

The quantitative results are summarized in Table~\ref{tab:scale_comparison}.

The results reveal a distinct dichotomy based on task nature:
\begin{itemize}
    \item \textbf{Reasoning-Intensive Tasks:} On challenging mathematical benchmarks, the 8B model with \textbf{DynaDebate} significantly outperforms the 32B baseline. Most notably on AIME 2025, which features difficult out-of-distribution problems, our method achieves a \textbf{13.34\%} improvement over the 32B model (36.67\% vs. 23.33\%). Similarly, on MATH500, the smaller model surpasses the larger one by 3.0\%. This suggests that for logic-heavy tasks, the quality of the reasoning process (enhanced by debate and verification) is more decisive than raw parameter count.

    \item \textbf{Knowledge-Intensive Tasks:} Conversely, on tasks requiring extensive world knowledge (MMLU) or factual retrieval (Biography), the 32B model retains an advantage. The Biography score, in particular, drops by 6.37\% with the smaller model. This aligns with the understanding that parametric knowledge scales with model size and cannot be fully compensated for by reasoning strategies alone.
\end{itemize}

These findings demonstrate that DynaDebate effectively activates the latent reasoning potential of smaller models. It allows an 8B parameter model to achieve state-of-the-art performance on complex reasoning tasks, rivaling or exceeding models 4$\times$ its size, offering a promising direction for deploying resource-efficient reasoning systems.

\section{Impact of Tool Augmentation on Single-Agent Baselines}
\label{sec:appendix_tool_ablation}

\subsection{Motivation and Setup}
A potential concern regarding the performance gains of DynaDebate is whether the improvements stem primarily from the integration of external tools rather than the proposed multi-agent debate mechanism itself. To rigorously disentangle the contribution of the debate framework from the benefits of tool utilization, we conducted additional comparative experiments using tool-augmented single-agent baselines.

\subsection{Implementation Details}
We established a strong baseline, denoted as Tool-Augmented CoT, which equips the standard Chain-of-Thought (CoT) method with external utilities. Crucially, to ensure a fair comparison, the tool availability for the baseline mirrors the configuration of the Verification Agent in our main experiments:

\begin{itemize}
    \item For MATH500: The baseline is equipped with a Python Code Interpreter (CoT + Code). It is prompted to write and execute Python code to perform calculations or verify logic steps.
    \item For MMLU: The baseline is equipped with a Search Engine (CoT + Search). It is allowed to query external information to retrieve factual knowledge before generating the final answer.
\end{itemize}

\subsection{Results}
The comparative results between the standard single-agent baseline (Standard CoT), the tool-augmented baseline (Tool-Augmented CoT), and our proposed DynaDebate framework are presented in Table~\ref{tab:tool_ablation}.

\begin{table}[h]
\centering
\small
\begin{tabular}{lcc}
\toprule
\textbf{Method} & \textbf{MATH500} & \textbf{MMLU} \\
\midrule
\multicolumn{3}{l}{\textit{\textbf{GPT-4o-mini}}} \\
CoT & 70.20 & 82.67 \\
\textit{Tool-Augmented CoT} & 70.60 & 83.00 \\
DynaDebate (Ours) & \textbf{73.60} & \textbf{83.33} \\
\midrule
\multicolumn{3}{l}{\textit{\textbf{Qwen3-8B}}} \\
Standard CoT & 78.00 & 78.67 \\
\textit{Tool-Augmented CoT} & 77.20 & 81.33 \\
DynaDebate (Ours) & \textbf{83.80} & \textbf{83.00} \\
\bottomrule
\end{tabular}
\caption{Performance comparison with tool-augmented single-agent baselines across distinct backbones.}
\label{tab:tool_ablation}
\end{table}

\noindent\textbf{Analysis of Results}

The results in Table~\ref{tab:tool_ablation}, considered alongside our main ablation studies (Table~\ref{tab:ablation}), provide a nuanced understanding of the relationship between tool availability and reasoning performance:

\begin{enumerate}
    \item \textbf{Tools are not a panacea; improper usage can hinder performance.}
    A critical observation from the Qwen3-8B results on MATH500 is that naively equipping a single agent with a code interpreter actually \textit{decreased} accuracy from 78.00\% to 77.20\%. This indicates that without a structured framework to verify and critique code execution, agents are susceptible to "tool misuse" (e.g., generating erroneous code or misinterpreting outputs), which introduces noise rather than clarity. DynaDebate avoids this by using tools strictly under triggered conditions within a peer-review setting, achieving 83.80\%.

    \item \textbf{Synergy between Search and Debate exceeds Search alone.}
    On the knowledge-intensive MMLU benchmark, we acknowledge that the Search Engine provides a tangible benefit to the single-agent baseline (e.g., Qwen3-8B improved from 78.67\% to 81.33\%). However, DynaDebate still achieves a superior performance (83.00\%). This demonstrates that our performance gains are not solely derived from accessing external information, but from the \textbf{DynaDebate framework's ability to effectively synthesize, debate, and verify} that information. The combination of search retrieval with multi-agent critique yields better outcomes than retrieval alone.

    \item \textbf{Architectural superiority persists without tools.}
    Finally, it is crucial to reference our ablation study in Section~\ref{sec:ablation} (Table~\ref{tab:ablation}). Even when the Verification module (and thus all tool access) is removed, the core DynaDebate framework (consisting only of Dynamic Path Generation and Debate) still outperforms the standard CoT baseline on benchmarks like MATH500 (81.80\% vs. 78.00\% for Qwen3-8B). This confirms that while tools act as a powerful amplifier, the fundamental reasoning improvements are driven by the proposed multi-agent mechanisms.
\end{enumerate}

\section{Computation Cost Details}
\label{app:cost}

Table~\ref{tab:performance_token} provides a round-by-round breakdown of accuracy and average token consumption for DynaDebate and SoM on MATH500. While DynaDebate introduces additional overhead due to path generation and multi-agent coordination, the overhead scales linearly with the number of debate rounds and remains moderate. Crucially, DynaDebate achieves its accuracy peak at Round 2, meaning the coordination cost is paid for only two rounds in the recommended configuration.

\begin{table}[h]
\centering
\begin{tabular}{c|cc|cc}
\hline
Round & \multicolumn{2}{c|}{Accuracy (\%)} & \multicolumn{2}{c}{Avg Tokens} \\
      & Ours & SoM & Ours & SoM \\
\hline
1 & 81.4 & \textbf{81.6} & 4779  & 3038  \\
2 & \textbf{83.8} & 79.0 & 18121 & 13406 \\
3 & 82.8 & 78.8 & 30134 & 21430 \\
4 & 82.8 & 80.8 & 42433 & 29600 \\
\hline
\end{tabular}
\caption{Round-by-round performance and average token consumption comparison between DynaDebate and SoM on MATH500.}
\label{tab:performance_token}
\end{table}

\section{Verification Trigger Frequency and Path Generation Statistics}
\label{app:trigger_stats}

\subsection{Verification Trigger Frequency}

Table~\ref{tab:trigger_rate} reports the average verification trigger rate across benchmarks and backbone models. A question is counted as ``triggered'' if at least one verification call is made during its debate.

\begin{table}[h]
\centering
\resizebox{\columnwidth}{!}{%
\begin{tabular}{lccc}
\toprule
\textbf{Benchmark} & \textbf{GPT-4o-mini} & \textbf{Qwen3-8B} & \textbf{Llama-3.3-70B} \\
\midrule
GSM8K       & 12.2\% & 12.1\% &  5.0\% \\
MMLU        & 25.5\% & 31.8\% & 17.2\% \\
MATH500     & 40.4\% & 29.3\% & 30.4\% \\
GPQA        & 53.7\% & 58.1\% & 35.0\% \\
AIME 2024   & 91.1\% & 84.4\% & 78.9\% \\
AIME 2025   & 92.2\% & 82.2\% & 88.9\% \\
\bottomrule
\end{tabular}
}
\caption{Average verification trigger rate (\%) per benchmark and backbone model. Benchmarks are ordered by increasing difficulty.}
\label{tab:trigger_rate}
\end{table}

The trigger rates exhibit a clear correlation with task difficulty: the verification agent fires on fewer than 13\% of GSM8K problems but on over 78\% of AIME problems across all backbones. This pattern confirms that the verification module is adaptive rather than indiscriminate---it allocates additional computational effort selectively to problems where agent disagreement genuinely arises. As a consequence, the token overhead reported in Appendix~\ref{app:cost} is concentrated on the hardest tasks, precisely where the accuracy gains from verification are largest.

\subsection{Path Generation Statistics}

Table~\ref{tab:path_gen} reports the average number of distinct reasoning paths generated per question by the Path Generation module ($\Phi_{\text{gen}}$). The target allocation is three paths (one per agent).

\begin{table}[h]
\centering
\resizebox{\columnwidth}{!}{%
\begin{tabular}{lccc}
\toprule
\textbf{Benchmark} & \textbf{GPT-4o-mini} & \textbf{Qwen3-8B} & \textbf{Llama-3.3-70B} \\
\midrule
GSM8K     & 3.00 & 2.98 & 2.84 \\
MMLU      & 3.00 & 3.00 & 3.00 \\
MATH500   & 2.95 & 2.92 & 2.33 \\
GPQA      & 3.00 & 3.00 & 3.00 \\
AIME 2024 & 2.91 & 3.00 & 2.69 \\
AIME 2025 & 3.00 & 3.00 & 2.62 \\
\bottomrule
\end{tabular}
}
\caption{Average number of reasoning paths generated per question. The target is 3 paths (one per agent).}
\label{tab:path_gen}
\end{table}

Path generation is highly reliable: GPT-4o-mini and Qwen3-8B consistently produce the full complement of three paths across all benchmarks. Llama-3.3-70B occasionally generates fewer paths on mathematical tasks (e.g., 2.33 on MATH500), reflecting variation in instruction-following capacity across backbone models; the framework gracefully falls back to the available paths in such cases without disrupting the debate pipeline.

\subsection{Path Independence: Single-path Rate vs.\ Path Coverage Rate}

To directly assess whether the generated paths are genuinely complementary rather than redundant paraphrases, we compare two metrics computed from each agent's \emph{pre-debate} initial answer (i.e., before any inter-agent interaction):

\begin{itemize}
    \item \textbf{Single-path Rate}: fraction of correct initial answers averaged over all paths and questions, equivalent to the accuracy of a single randomly selected path.
    \item \textbf{Path Coverage Rate}: fraction of questions for which \emph{at least one} of the three paths yields a correct initial answer.
\end{itemize}

If the paths were near-identical, coverage would barely exceed the single-path rate. A large gap instead indicates that different paths succeed on different questions---the defining signature of genuine complementarity.

\begin{table}[h]
\centering
\resizebox{\columnwidth}{!}{%
\begin{tabular}{llcc}
\toprule
\textbf{Benchmark} & \textbf{Model} & \textbf{Single-path} & \textbf{Coverage} \\
\midrule
{GSM8K}
 & GPT-4o-mini    & 91.1\% & 95.3\% \\
 & Qwen3-8B       & 90.8\% & 95.2\% \\
 & Llama-3.3-70B  & 95.0\% & 96.0\% \\
\midrule
{MMLU}
 & GPT-4o-mini    & 79.4\% & 89.3\% \\
 & Qwen3-8B       & 76.9\% & 88.2\% \\
 & Llama-3.3-70B  & 84.8\% & 91.9\% \\
\midrule
{MATH500}
 & GPT-4o-mini    & 66.7\% & 77.4\% \\
 & Qwen3-8B       & 75.8\% & 84.6\% \\
 & Llama-3.3-70B  & 69.3\% & 76.4\% \\
\midrule
{GPQA}
 & GPT-4o-mini    & 43.6\% & 64.0\% \\
 & Qwen3-8B       & 43.5\% & 64.0\% \\
 & Llama-3.3-70B  & 58.4\% & 74.8\% \\
\midrule
{AIME 2024}
 & GPT-4o-mini    &  6.7\% & 11.1\% \\
 & Qwen3-8B       & 13.7\% & 21.1\% \\
 & Llama-3.3-70B  & 14.1\% & 20.0\% \\
\midrule
{AIME 2025}
 & GPT-4o-mini    &  8.1\% & 13.3\% \\
 & Qwen3-8B       & 20.7\% & 32.2\% \\
 & Llama-3.3-70B  &  5.2\% & 12.2\% \\
\bottomrule
\end{tabular}
}
\caption{Single-path Rate vs.\ Path Coverage Rate (\%) measured from pre-debate initial answers. Benchmarks are ordered by increasing difficulty.}
\label{tab:path_coverage}
\end{table}

As shown in Table~\ref{tab:path_coverage}, Path Coverage Rate consistently and substantially exceeds Single-path Rate across all benchmarks and backbones. The gap is most pronounced on GPQA ($+$16--20 percentage points), where expert-domain problems admit only a narrow set of valid reasoning approaches and paths can succeed or fail independently. Moderate but consistent gaps also appear on MATH500 ($+$7--9 pp) and MMLU ($+$7--11 pp). Even on the hardest AIME tasks, where absolute accuracy is low, coverage exceeds single-path rate by 5--12 pp, confirming that the three paths explore genuinely distinct solution trajectories rather than converging on the same (often incorrect) approach. The relatively small gap on GSM8K ($+$1--4 pp) is expected: when problems are sufficiently easy, most paths independently reach the correct answer, leaving little room for complementary coverage. Together, these results confirm that $\Phi_{\text{gen}}$ produces paths that are functionally independent, providing the debate with a diverse pool of reasoning candidates from which the best can be selected and refined.

\section{Prompts}
\label{app:prompts}
In this section, we provide the specific prompts used in the DynaDebate framework. To provide a concrete demonstration of the instruction design, we use the MATH500 dataset as an illustrative example throughout this section.

\subsection{Path Generation and Allocation}
\label{app:prompt_path}

\begin{tcolorbox}[
    breakable,
    colback=white,
    colframe=black,
    boxrule=0.8pt,
    left=6pt,
    right=6pt,
    top=6pt,
    bottom=6pt
]
\ttfamily\small
System Instruction:\\
You are participating in a collaborative problem-solving session for a challenging math competition problem.

\vspace{0.5em}
--- STRATEGIC BRAINSTORMING PHASE ---\\
Your task is to identify genuinely independent and feasible mathematical approaches that could solve the problem.
A ``path'' here refers to a distinct mathematical strategy, such as a specific logical approach, theorem application,
or conceptual framework (e.g., algebraic vs.\ geometric, recursion vs.\ combinatorics).

\vspace{0.5em}
Problem: \lbrack Problem Description\rbrack

\vspace{0.5em}
Your responsibilities:

\vspace{0.25em}
PART 1: METHOD EXPLORATION\\
1. List up to 3 viable candidate strategies that could be used to solve the question.\\
2. List ONLY genuinely distinct strategies.
\begin{itemize}
    \item If only 1 or 2 viable strategies exist, list only those.
    \item Do NOT invent additional strategies merely to increase the count.
    \item Strategies that differ only in notation or minor algebraic manipulation but rely on the same core theorem
    do NOT count as independent.
\end{itemize}

\vspace{0.25em}
PART 2: RIGOROUS FEASIBILITY ASSESSMENT\\
For each strategy, specify:
\begin{itemize}
    \item Core idea (key theorem, formula, or logical technique)
    \item Expected reliability (High / Medium / Low)
\end{itemize}

\vspace{0.25em}
PART 3: COLLABORATION STRATEGY RECOMMENDATION\\
Output strictly in the following format:

\begin{verbatim}
"METHOD_1:"
- Core idea: ...
- Critical step: ...

"METHOD_2:"
...
\end{verbatim}
\end{tcolorbox}

\subsection{Process-Centric Debate}
\label{app:prompt_debate}

\begin{tcolorbox}[
    breakable,
    colback=white,
    colframe=black,
    boxrule=0.8pt,
    left=6pt,
    right=6pt,
    top=6pt,
    bottom=6pt
]
\ttfamily\small
System Instruction:\\
Please solve the following high school competition math problem. Provide a step-by-step reasoning for your solution.
Use LaTeX for all mathematical expressions.

\vspace{0.5em}
Problem: \lbrack Problem Description\rbrack

\vspace{0.5em}
--- COMPARISON PHASE (Round N) ---

\vspace{0.25em}
Your Original Method: \lbrack Method Name\rbrack: \lbrack Method Description\rbrack\\
You are Agent \lbrack ID\rbrack.

\vspace{0.25em}
Previous Round Results from All Agents:\\
Agent 1 Result: [Response 1]\\
...

\vspace{0.25em}
[Optional: Code Verification Feedback Inserted Here]

\vspace{0.5em}
--- YOUR TASK ---

\vspace{0.25em}
Objective: Compare results, consider verification feedback, and refine your solution.

\vspace{0.25em}
Analysis Required:\\
1. Critically review the reasoning from other agents. Do you see any logical flaws or calculation mistakes?\\
2. Consider the computational verification results. Re-evaluate your own previous reasoning in light of their approaches.\\
3. Re-evaluate your own previous reasoning in light of their approaches.

\vspace{0.5em}
Given a mathematics problem, determine the answer through comparison and verification feedback.
Your final answer should be in the form \verb|\boxed{answer}|, at the end of your response.
\end{tcolorbox}
\label{fig:prompt_debate}

\subsection{Verification Mechanism}
\label{app:prompt_verification}

\begin{tcolorbox}[
    breakable,
    colback=white,
    colframe=black,
    boxrule=0.8pt,
    left=6pt,
    right=6pt,
    top=6pt,
    bottom=6pt
]
\ttfamily\small
System Instruction:\\
You are in a multi-agent collaborative setting for solving a math problem, and your task is to generate code to verify the answers.

\vspace{0.5em}
--- PROBLEM --- \\
\lbrack Problem Description\rbrack

\vspace{0.5em}
--- VERIFICATION TRIGGER --- \\
Reason for verification: \lbrack Trigger Reason\rbrack

\vspace{0.5em}
--- RESPONSES TO VERIFY --- \\
Agent 1 Response: ...\\
Agent 2 Response: ...

\vspace{0.5em}
--- YOUR TASK ---

\vspace{0.25em}
Objective: Use computational verification to analyze the responses and provide feedback for the next round of debate.

\vspace{0.25em}
Verification Strategy:\\
1. Extract Key Claims: Identify the main mathematical claims and final answers.\\
2. Design Verification Code: Write Python code to solve the current problem.\\
3. Execute and Compare: Run your code and compare results with agent responses.\\
4. Identify Issues: Determine which responses (if any) contain errors.\\
5. Provide Constructive Feedback: Give specific guidance for the next round.

\vspace{0.5em}
CRITICAL REQUIREMENTS FOR CODE:
\begin{itemize}
    \item Store your final result in a variable named `ans'
    \item Write clean, readable Python code
    \item Use appropriate mathematical libraries (math, numpy, sympy, etc.)
\end{itemize}

\vspace{0.25em}
Your response must include:\\
1. Analysis Summary: Brief explanation of discrepancies or concerns found.\\
2. Verification Code: Python code in python blocks.\\
3. Agent-Specific Feedback: Specific feedback for each agent.\\
4. Recommendations: Concrete suggestions for improving solutions.

\vspace{0.5em}
Format your code properly. This feedback will be provided to all agents in the next round.
\end{tcolorbox}
\label{fig:prompt_verification}

\section{Case Study}
\label{app:case_study}

To qualitatively demonstrate the efficacy of DynaDebate, we analyze a representative failure case from the MATH500 dataset.

\begin{tcolorbox}[
    colback=white,
    colframe=black!70,
    boxrule=0.8pt,
    title=\textbf{Problem Example (MATH500)},
    fonttitle=\small\sffamily\bfseries,
    sharp corners,
    bottom=5pt
]
    \small
    \textbf{Question:} The expression $2\cdot 3 \cdot 4\cdot 5+1$ is equal to 121... In total, how many values can be obtained from the expression $2\cdot 3\cdot 4 \cdot 5 + 1$ by inserting parentheses? (Note that rearranging terms is not allowed, only inserting parentheses).

    \vspace{0.2cm}
    \textbf{Ground Truth:} 4 values ($144, 126, 122, 121$).
    \begin{itemize}[nosep, leftmargin=1em]
        \item $2\cdot 3 \cdot 4 \cdot (5 + 1) = 144$
        \item $2\cdot 3 \cdot (4 \cdot 5 + 1) = 126$
        \item $2\cdot (3 \cdot 4 \cdot 5 + 1) = 122$  \textcolor{red}{\textbf{(Often missed by baselines)}}
        \item $(2\cdot 3 \cdot 4 \cdot 5) + 1 = 121$
    \end{itemize}
\end{tcolorbox}

\begin{tcolorbox}[
    breakable,
    colback=red!5!white,
    colframe=red!60!black,
    title=\textbf{\sffamily Baseline (SoM)},
    fonttitle=\small\bfseries,
    boxrule=0.5pt,
    sharp corners,
    left=2pt, right=2pt, top=2pt, bottom=2pt
]
    \footnotesize
    \textbf{Round 1:}
    \begin{itemize}[leftmargin=1.5em, nosep]
        \item \textbf{Agent A:} The expression is $2 \cdot 3 \cdot 4 \cdot 5 + 1$. By operator precedence, it evaluates to $120 + 1 = 121$.

Multiplication is associative, so any parentheses applied only to the product $2\cdot3\cdot4\cdot5$ still yield 120. Thus, all such groupings give 121.

However, if we group the addition early, e.g. $(5+1)$, new values arise:
\[
(2 \cdot 3) \cdot (4 \cdot (5 + 1)) = 6 \cdot 24 = 144.
\]
Enumerating all 14 valid parenthesizations shows only two distinct results: **121** and **144**.

\medskip
\noindent\textbf{Final Answer:} $\boxed{2}$

        \item \textbf{Agent B:} The base value is $(2 \cdot 3 \cdot 4 \cdot 5) + 1 = 121$.

Since multiplication is associative and commutative, re-grouping the product alone does not change the result.

Only when parentheses include the ``$+1$'' in a multiplicative subexpression do we get a new value:
\[
(2 \cdot (3 \cdot 4)) \cdot (5 + 1) = 24 \cdot 6 = 144.
\]
All other valid groupings either yield 121 or 144. No third value appears.

\medskip
\noindent\textbf{Final Answer:} $\boxed{2}$

        \item \textbf{Agent C:}We consider groupings where addition occurs at different depths:

\begin{align*}
(2 \cdot 3 \cdot 4 \cdot 5) + 1 &= 121, \\
(2 \cdot (3 \cdot 4)) \cdot (5 + 1) &= 144, \\
2 \cdot (3 \cdot ((4 \cdot 5) + 1)) &= 2 \cdot (3 \cdot 21) = 126.
\end{align*}

Other groupings replicate these values. Thus, three distinct results are found: 121, 144, and 126.

\medskip
\noindent\textbf{Final Answer:} $\boxed{3}$

    \end{itemize}

    \vspace{0.1cm}
    \textbf{Round 2 :}
    \begin{itemize}[leftmargin=1.5em, nosep]
        \item \textbf{Agent A :} The expression $2 \cdot 3 \cdot 4 \cdot 5 + 1$ evaluates to 121 by default.

There are 14 valid parenthesizations (4th Catalan number). Evaluating them yields:

\begin{align*}
\text{Most groupings} &\Rightarrow 121, \\
\text{Groupings with } (5+1) &\Rightarrow 144, \\
2 \cdot (3 \cdot ((4 \cdot 5) + 1)) &= 2 \cdot (3 \cdot 21) = 126.
\end{align*}

Thus, the distinct values are 121, 144, and 126.

\medskip
\noindent\textbf{Final Answer:} $\boxed{3}$

        \item \textbf{Agent B :} Although there are 42 parenthesizations (5th Catalan number), associativity of multiplication collapses most to the same value.

Key distinct evaluations:
\begin{align*}
(2 \cdot 3 \cdot 4 \cdot 5) + 1 &= 121, \\
(2 \cdot (3 \cdot 4)) \cdot (5 + 1) &= 144, \\
2 \cdot (3 \cdot ((4 \cdot 5) + 1)) &= 126.
\end{align*}

All other groupings reproduce one of these three values.

\medskip
\noindent\textbf{Final Answer:} $\boxed{3}$

        \item \textbf{Agent C:} Using recursive evaluation of all valid parenthesizations of $2 \cdot 3 \cdot 4 \cdot 5 + 1$, we find only three distinct results:

\[
121,\quad 144,\quad 126,
\]
corresponding to:
\begin{itemize}[left=0pt,topsep=2pt,itemsep=0pt]
    \item addition applied last,
    \item $5+1$ grouped early,
    \item $(4\cdot5 + 1)$ grouped before outer multiplications.
\end{itemize}

No other values arise under the no-rearrangement constraint.

\medskip
\noindent\textbf{Final Answer:} $\boxed{3}$
    \end{itemize}

    \vspace{0.1cm}
    \textbf{VOTED Answer:} $\boxed{3}$ \textcolor{red}{Incorrect}
\end{tcolorbox}

\begin{tcolorbox}[
    breakable,
    lines before break=0,
    enlarge top initially by=0mm,
    pad at break=1mm,
    colback=blue!5!white,
    colframe=blue!60!black,
    title=\textbf{\sffamily DynaDebate (Ours)},
    fonttitle=\small\bfseries,
    boxrule=0.5pt,
    sharp corners,
    left=2pt, right=2pt, top=2pt, bottom=2pt
]
    \footnotesize
    \textbf{Phase 1: Path Allocation}
    \begin{itemize}[leftmargin=1.5em, nosep]
        \item \textbf{Path Generation Agent :}Three distinct solution strategies:

\begin{enumerate}[left=0pt,topsep=2pt,itemsep=0pt]
    \item \textbf{Combinatorics with Parsing Trees}: Model parenthesizations as binary trees; high reliability.
    \item \textbf{Dynamic Programming with Memoization}: Recursively compute all sub-expression values; high reliability.
    \item \textbf{Operator Precedence Analysis}: Manually inspect groupings; medium reliability for small expressions.
\end{enumerate}

    \end{itemize}

    \vspace{0.1cm}
    \textbf{Phase 2: Debate Round 1 }
    \begin{itemize}[leftmargin=1.5em, nosep]
        \item \textbf{Agent A:} Evaluating all 14 valid parenthesizations of $2 \cdot 3 \cdot 4 \cdot 5 + 1$ yields four distinct values:

\begin{align*}
(2\cdot3\cdot4\cdot5) + 1 &= 121, \\
2 \cdot (3 \cdot ((4 \cdot 5) + 1)) &= 126, \\
2 \cdot ((3 \cdot (4 \cdot 5)) + 1) &= 122, \\
\text{Any grouping with } (5+1) \text{ in product} &= 144.
\end{align*}

Thus, the number of distinct values is $\boxed{4}$.
        \item \textbf{Agent B:} Using dynamic programming on tokens $[2,\cdot,3,\cdot,4,\cdot,5,+,1]$, we compute:

\[
\texttt{dp}[0][4] = \{121,\ 144,\ 126\}.
\]

The split at the final '+' gives 121; splits before '+' with $(5+1)$ yield 144 or 126.
No 122 is generated due to state aggregation.

\medskip
\noindent\textbf{Final Answer:} $\boxed{3}$
        \item \textbf{Agent C:} Two cases only:
\begin{itemize}[left=0pt,topsep=2pt,itemsep=0pt]
    \item Addition last: $(2\cdot3\cdot4\cdot5)+1 = 121$,
    \item Addition grouped early (e.g., $5+1$): yields $144$.
\end{itemize}

All other groupings reproduce one of these. Thus, only two distinct values exist.

\medskip
\noindent\textbf{Final Answer:} $\boxed{2}$
    \end{itemize}

    \vspace{0.1cm}
    \textbf{Phase 3: Debate Round 2 }
    \begin{itemize}[leftmargin=1.5em, nosep]
        \item \textbf{Agent A:} Systematic evaluation of all valid parenthesizations of $2 \cdot 3 \cdot 4 \cdot 5 + 1$ yields four distinct values:

\begin{align*}
(2\cdot3\cdot4\cdot5) + 1 &= 121, \\
2 \cdot (3 \cdot ((4 \cdot 5) + 1)) &= 126, \\
2 \cdot ((3 \cdot (4 \cdot 5)) + 1) &= 122, \\
\text{Any grouping with } (5+1) \text{ in product} &= 144.
\end{align*}

Thus, the total number of distinct values is $\boxed{4}$.
        \item \textbf{Agent B :} Despite using dynamic programming, the evaluation only considered two outcomes:
\begin{itemize}[left=0pt,topsep=2pt,itemsep=0pt]
    \item Standard precedence: $121$,
    \item Grouping with $(5+1)$: $144$.
\end{itemize}
Other valid groupings (e.g., $2 \cdot (3 \cdot (20 + 1)) = 126$) were missed due to incomplete state enumeration.

\medskip
\noindent\textbf{Final Answer:} $\boxed{2}$
        \item \textbf{Agent C:} A complete dynamic programming evaluation over tokens $[2,\cdot,3,\cdot,4,\cdot,5,+,1]$ yields the full set of possible values:

\[
\{121,\ 144,\ 126,\ 122\}.
\]

These correspond to:
\begin{itemize}[left=0pt,topsep=2pt,itemsep=0pt]
    \item addition last,
    \item $(5+1)$ grouped early,
    \item $(4\cdot5 + 1)$ grouped,
    \item $(3\cdot4\cdot5 + 1)$ grouped.
\end{itemize}

All four are valid and distinct under the no-rearrangement constraint.

\medskip
\noindent\textbf{Final Answer:} $\boxed{4}$
    \end{itemize}

    \vspace{0.1cm}
    \textbf{VOTED Answer:} $\boxed{4}$ \textcolor{green}{Correct}

\end{tcolorbox}

\captionof{figure}{Comparison on a MATH500 problem involving operator precedence.In SoM, all agents overlooked the answer 122, whereas in DynaDebate, by assigning diverse reasoning paths, the agents obtained the correct answer (4); through structured debate, the influence of the correct answer was amplified and ultimately confirmed via voting.}
\label{fig:case_study_math}

\end{document}